\documentclass[lettersize,journal]{IEEEtran}
\usepackage{amsmath,amsfonts}
\usepackage{algorithmic}
\usepackage{array}
\usepackage[caption=false,font=normalsize,labelfont=sf,textfont=sf]{subfig}
\usepackage{textcomp}
\usepackage{stfloats}
\usepackage{url}
\usepackage{verbatim}
\usepackage{graphicx}
\usepackage{cite}
\usepackage{picture}
\usepackage{verbatim}
\usepackage{pgfplots}
\usepackage{booktabs}
\usepackage{multicol}
\usepackage{multirow}
\usepackage{xcolor}
\usepackage{amssymb}
\usepackage{caption}
\usepackage{array}
\usepackage{float}
\usepackage{subcaption}
\usepackage{bm}

\newcommand\etal{\emph{et al. }}
\newcommand\red{\textcolor{red}}
\newcommand\blue{\textcolor{blue}}

\begin{document}
\title{All-in-one Weather-degraded Image Restoration via Adaptive Degradation-aware Self-prompting Model}

\author{Yuanbo Wen, Tao Gao, Ziqi Li, Jing Zhang, Kaihao Zhang, Ting Chen
\thanks{Manuscript received January 25, 2024; revised July 12, 2024; accepted October 4, 2024.}
\thanks{Yuanbo~Wen, Ziqi~Li and Ting~Chen are with the School of Information Engineering, Chang'an University, Xi'an, China. E-mail: \{wyb@chd.edu.cn; lzq@chd.edu.cn; tchenchd@126.com\}}
\thanks{Tao~Gao is with the School of Data Science and Artificial Intelligence, Chang'an University, Xi'an, China. E-mail: \{tgaochd@126.com\}}
\thanks{Jing~Zhang is with the School of Computing, Australian National University, Canberra, ACT, Australia. E-mail: \{jing.zhang@anu.edu.au\}}
\thanks{Kaihao~Zhang is with the School of Computer Science and Technology, Harbin Institute of Technology, Shenzhen, China. E-mail: \{super.khzhang@gmail.com;\}}
}

\markboth{IEEE Transactions on Multimedia}
{All-in-one Weather-degraded Image Restoration via Adaptive Degradation-aware Self-prompting Model}

\maketitle

\begin{abstract}

Existing approaches for all-in-one weather-degraded image restoration suffer from inefficiencies in leveraging degradation-aware priors, resulting in sub-optimal performance in adapting to different weather conditions.
To this end, we develop an adaptive degradation-aware self-prompting model (ADSM) for all-in-one weather-degraded image restoration.
Specifically, our model employs the contrastive language-image pre-training model (CLIP) to facilitate the training of our proposed latent prompt generators (LPGs), which represent three types of latent prompts to characterize the degradation type, degradation property and image caption.
Moreover, we integrate the acquired degradation-aware prompts into the time embedding of diffusion model to improve degradation perception.
Meanwhile, we employ the latent caption prompt to guide the reverse sampling process using the cross-attention mechanism, thereby guiding the accurate image reconstruction.
Furthermore, to accelerate the reverse sampling procedure of diffusion model and address the limitations of frequency perception, we introduce a wavelet-oriented noise estimating network (WNE-Net).
Extensive experiments conducted on eight publicly available datasets demonstrate the effectiveness of our proposed approach in both task-specific and all-in-one applications.
\end{abstract}


\section{Introduction}
\label{sec:introduction}

\IEEEPARstart{R}{estoring} images degraded by adverse weather degradations represents significance within the domain of intelligent transportation \cite{pereira2024weather, gao2023novel, li2020all, wen2024encoder}, where the weather degradations can be rain, haze, raindrop, snow, low-lightness, rain-by-snow, rain-by-haze, rain-by-raindrop and other degradations.
Existing methods for weather-degraded image restoration can be categorized into task-specific, task-aligned and all-in-one settings.
The task-specific methods \cite{wen2024heavy, wen2023encoder, song2023vision, 10145603, wen2024neural} can only eliminate one type of degradations after the training, necessitating different models and parameters.
The task-aligned ones \cite{zamir2022restormer, wang2022uformer, wen2024unpaired, wen2023multi} can also remove one type of degradations, using the same model with varying parameters.
Both strategies exhibit limitations in adapting to diverse weather conditions.
In contrast, all-in-one methods \cite{li2020all, valanarasu2022transweather} accommodate various weather degradations using a single model with consistent parameters.

\begin{figure}[!t]
    \centering
    \includegraphics[width=\linewidth]{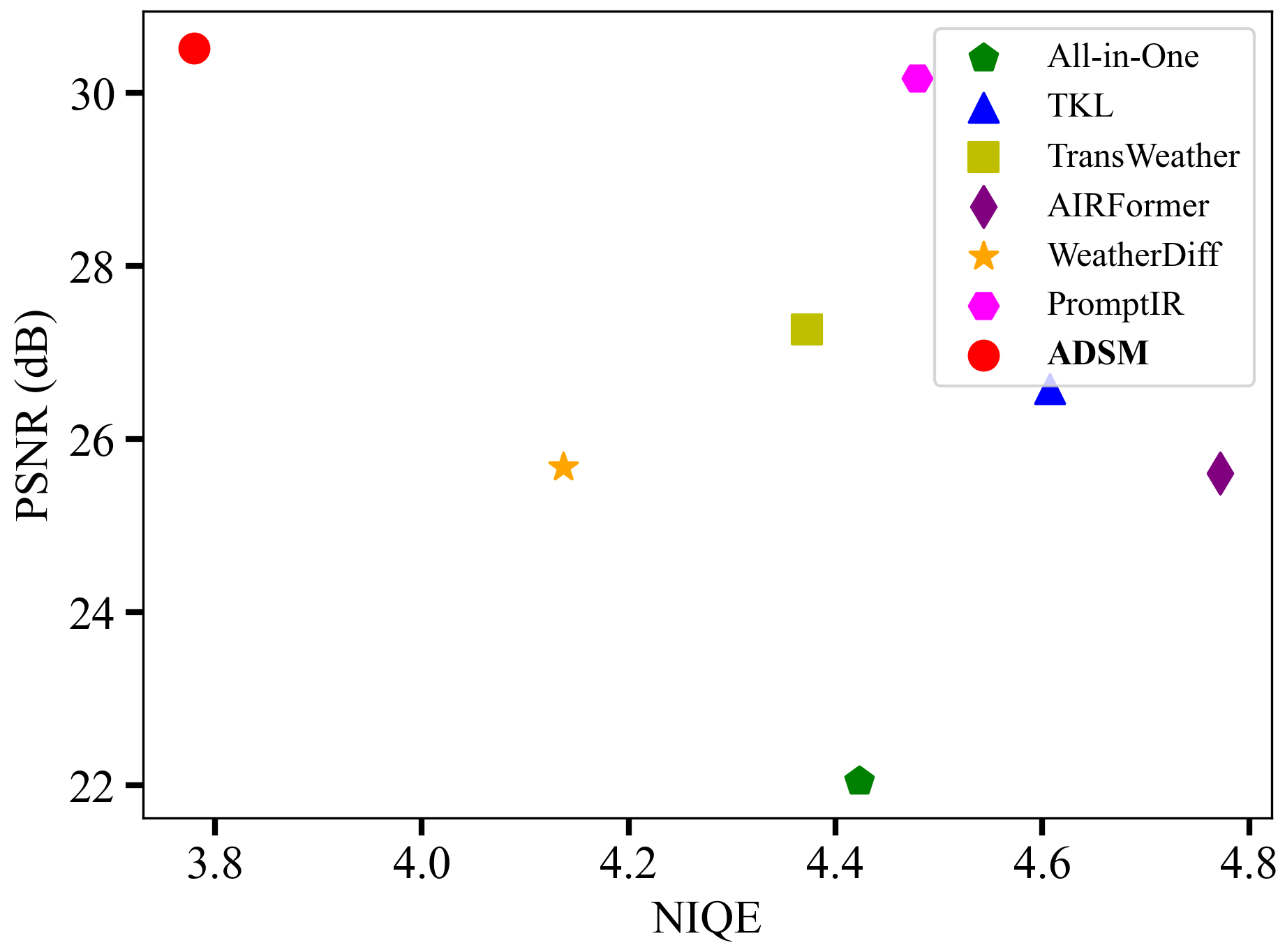}
    \caption{Intuitive comparisons of the existing all-in-one approaches and our proposed method between the supervised and un-supervised metrical scores of reconstructed images. Our model notably enhances the naturalness and image quality of generated images, surpassing the capabilities of existing methods in both aspects.}
    \label{fig:naturalness}
\end{figure}

\begin{figure*}[h]
    \centering
    \includegraphics[width=\linewidth]{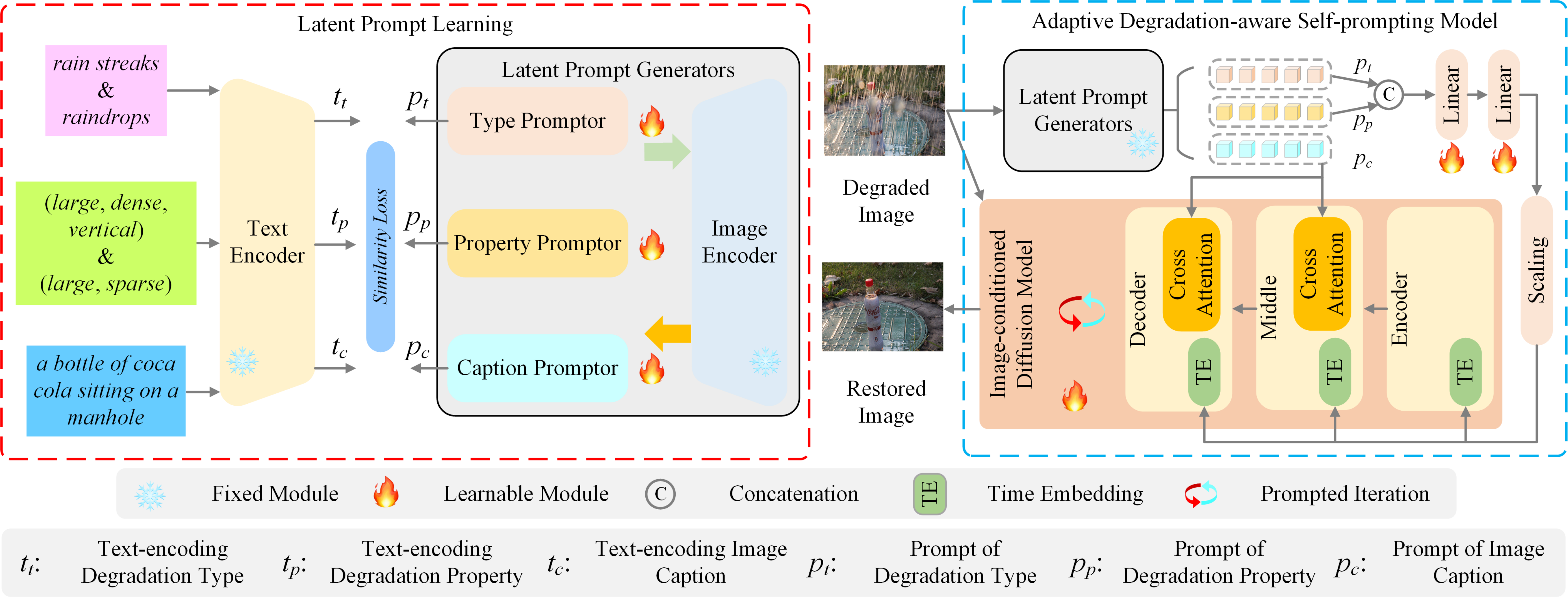}
    \caption{Overview of our proposed adaptive degradation-aware self-prompting model (ADSM) for all-in-one weather-degraded image restoration. Initially, we train three latent prompt generators (LPGs) to create the corresponding prompts for degradation type, degradation property and image caption. We then integrate these latent prompts into the diffusion model to perform iterative prompt learning, facilitating the weather degradation elimination and image content reconstruction.}
    \label{fig:overview}
\end{figure*}

Recently, there has been a growing interest in integrating the prompt learning methodologies to improve the perceptual capabilities of image restoration models for addressing specific weather degradations.
For instance, MAE-VQGAN \cite{bar2022visual} strategically utilizes the grid-like configurations during the inference phase to perform effective inpainting within regions lacking visual content.
Similarly, Painter \cite{wang2023images} generates desired images based on specified paired instance images.
Given the nature of conditional generation, solutions that rely on discrete prompt engineering \cite{liu2023pre, haviv2021bertese, wallace2019universal} are sensitive to the quality of given fixed prompts.
To address this limitation, ProRes \cite{ma2023prores} introduces the trainable degradation-aware visual prompts designed to model specific types of weather degradations. However, it also relies heavily on the high-quality learnable visual prompts and lacks the information related to specific degradation and image content, resulting in limited effectiveness for processing complex weather-degraded images.
Therefore, there is a need for prompt learning approachees that integrates additional prompts to characterize both the degradation specifics and contextual details.

In this work, we propose an effective adaptive degradation-aware self-prompting model (ADSM) for all-in-one weather-degraded image restoration (as shown in Figure \ref{fig:overview}), which enhances the naturalness of the reconstructed images while achieving the improved supervised metrical scores.
Our approach leverages the extensive vision-language priors in CLIP model \cite{radford2021learning, luo2023controlling}, enabling us to generate latent prompts specifically tailored to the weather-degraded images being processed.
To align the latent features of weather-degraded images with their corresponding ground truths, we train three latent prompt generators (LPGs), which are designed to produce the latent prompts of degradation type, degradation property and image caption, respectively.
By using the pre-trained and learnable image encoders in CLIP, our LPGs align the generated latent prompts with the output latent features from the text encoder in CLIP space.
LPGs can also help solve the issue where texts with similar meanings may appear significantly distant in CLIP space \cite{liang2023iterative}.
In our proposed LPGs, the degradation prompts for type and property identify the potential features of both individual and combined weather conditions, whereas the existing similar method \cite{luo2023controlling} fails to address combined weather degradations and to classify the specific degradation properties.
Additionally, the caption prompt is essential for verifying the clarity of content in reconstructed images.
After setting up the latent prompt generators, we utilize the image-conditioned diffusion model \cite{ho2020denoising, luo2023image} as the core restoration framework.
Here, we combine the prompts indicating degradation type and property, which are then incorporated into the time embedding, improving the degradation perception.
Meanwhile, we introduce the cross-attention mechanism to blend the latent caption prompt into the diffusion model, guiding the reconstruction of image content.

To improve the computational efficiency in sampling generated images and enhance the accuracy of noise estimation, we devise a wavelet-oriented noise estimating network (WNE-Net).
The traditional residual block in diffusion model \cite{ho2020denoising} struggles to fully capture the crucial global information needed for effective noise perception \cite{phung2023wavelet}.
In our WNE-Net, we introduce a wavelet self-attention representation block (WSRB) designed to effectively separate out different frequency components and capture a deep understanding of long-range relationships.
Meanwhile, the inherent down-sampling feature of discrete wavelet transform (DWT) \cite{mallat1989theory} reduces the computational burden.
In addition, we utilize the frequency partitioning in cross-stage sampling operations and introduce a wavelet feature sampling block (WFSB) that includes both down-sampling and up-sampling operators. WFSB utilizes two branches to capture the sampling features that are sensitive to frequency, refining these characteristics by identifying the specific degradation properties.
Our WNE-Net brings two key advancements. Firstly, it allows our model to better distinguish the inherent frequency characteristics of input signals. Secondly, it utilizes wavelet transform to adjust the feature resolutions, significantly reducing the computational load and accelerating the reverse sampling process.

Figure \ref{fig:overview} illustrates the overview of our proposed model. We summarize the main contributions as follows. 

\begin{itemize}
    \item We achieve all-in-one weather-degraded image restoration via an adaptive degradation-aware self-prompting model, which is prompted by the latent prompts of degradation type, degradation property and image caption.
    \item We utilize the latent degradation prompts of type and property to improve the understanding of weather conditions. Meanwhile, we employ the cross-attention mechanism to integrate the caption prompt into diffusion model, enhancing the accuracy of image reconstruction.
    \item We develop a wavelet-oriented noise estimating network to leverage the frequency separation principles for efficient and accurate noise estimation.
\end{itemize}

\section{Related Work}
\label{sec:related}

\subsection{All-in-one Weather Removal}
Previous approaches either create dedicated physical models for specific weather removal tasks \cite{10243133, peng2023u, wang2017single}, or develop unified solutions trained individually \cite{zamir2021multi, zamir2022restormer, wang2022uformer, 9369906}, without exploring the shared aspects among various weather degradations needed for comprehensive weather-degraded image restoration.
Recently, Li \etal \cite{li2020all} delineate a pioneering approach that employs the task-specific optimized encoders to address different weather degradations.
Subsequently, Valanarasu \etal \cite{valanarasu2022transweather} employ a spatially constrained self-attention mechanism together with task-related queries to simultaneously reduce various weather degradations.
Meanwhile, Chen \etal \cite{chen2022learning} utilize a two-phase method to blend knowledge, combined with a multi-contrastive learning approach.
Based on the diffusion model \cite{ho2020denoising}, $\ddot{\rm O}$zdenizc \etal \cite{ozdenizci2023restoring} present a patch-based diffusion model to recover high-resolution weather-degraded images, while Gao \etal \cite{gao2023frequency} explore the frequency variations during the weather removal procedure and propose a frequency-oriented all-in-one weather-degraded image restoration approach.
Moreover, Ye \etal \cite{ye2023adverse} achieve a cohesive strategy to reduce damage caused by adverse weather conditions using predefined guidelines.
However, these methods only capture weather degradation representations through basic data-driven self-training, without incorporating prompts to enhance the adaptability of all-in-one models when dealing with various types of weather degradation.

\subsection{Visual Prompt Learning}
Current research aims to utilize prompt learning techniques in practical visual applications. For instance, MAE-VQGAN \cite{bar2022visual} uses a grid-like input setup during inference to autonomously fill in missing visual content through inpainting operations. CoCoOp \cite{zhou2022conditional} introduces a new method integrating prompt acquisition into the adaptation of vision-language models for downstream visual tasks, improving the generalizability by conditioning prompts on specific input instances. ControlNet \cite{zhang2023adding} recently incorporates the controls into stable diffusion model to customize them for specific task requirements. Painter \cite{wang2023images} aligns network outputs with task-specific goals using paired visual prompts, while ProRes \cite{ma2023prores} employs the degradation-aware learnable prompts to guide image reconstruction. However, these methods heavily rely on high-quality paired or learnable visual prompts, resulting in dissimilar visual representations across different prompt settings in the final reconstructed images.

\section{Methodology}
\label{sec:methodology}

\subsection{Latent Prompt Generators}

\begin{figure}[h]
    \centering
    \includegraphics[width=\linewidth]{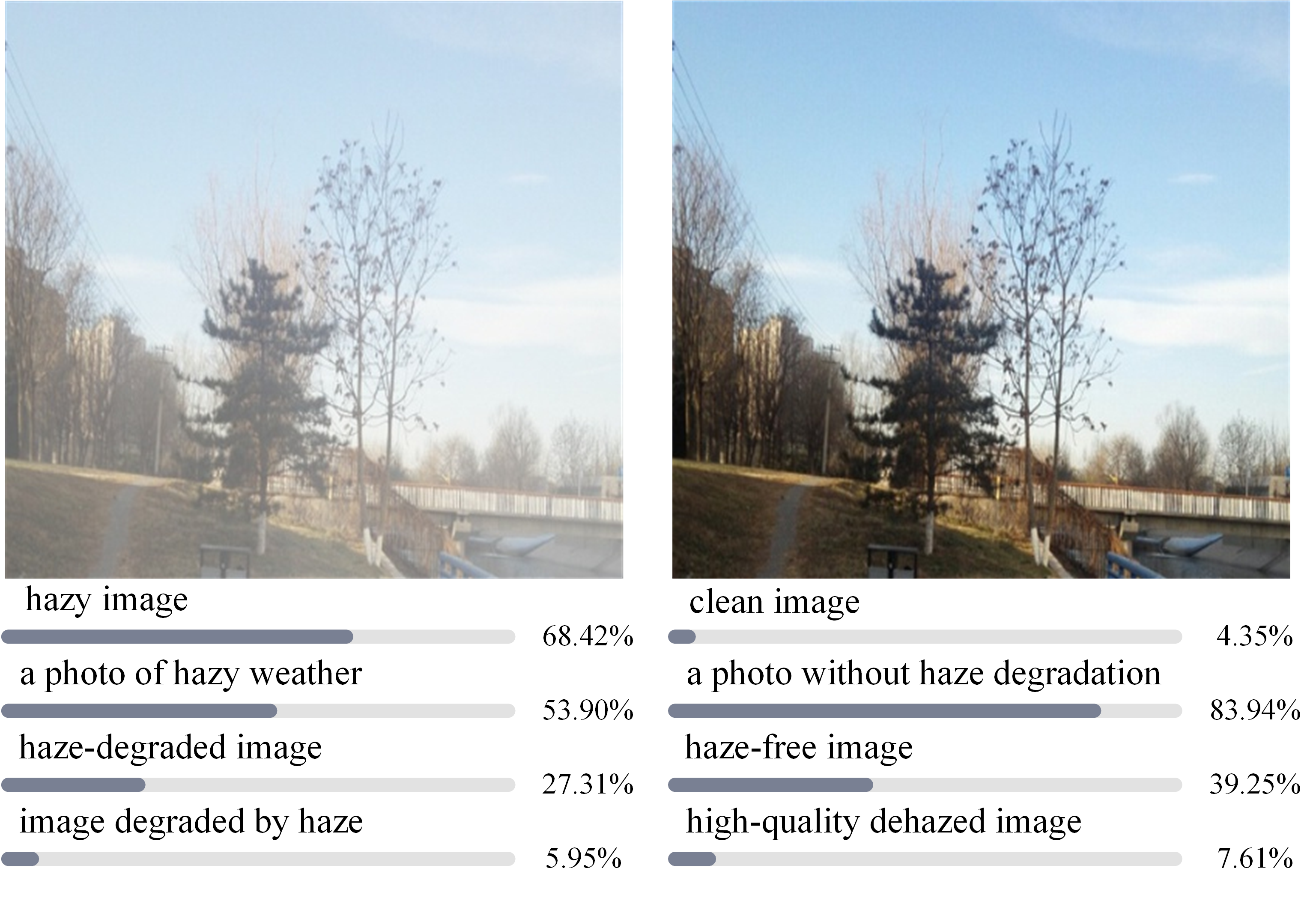}
    \caption{Two examples of texts that share similar meanings and depict the same image in the latent space. Despite their similar meanings, their latent encoding features differ.}
    \label{fig:prompts}
\end{figure}

In CLIP space, texts with similar meanings can exhibit significant distances \cite{liang2023iterative}, as illustrated in Figure \ref{fig:prompts}. Therefore, using texts to constrain the reverse sampling process leads to instability in the reconstructed images. To this end, we utilize both fixed and learnable image encoders of CLIP \cite{radford2021learning} as the core network for our latent prompt generators. Following \cite{luo2023controlling}, features from various levels of the unlocked encoder are combined with those from the locked encoder using convolution operations initialized with zero weights.

\begin{figure}[h]
     \centering
     \includegraphics[width=\linewidth]{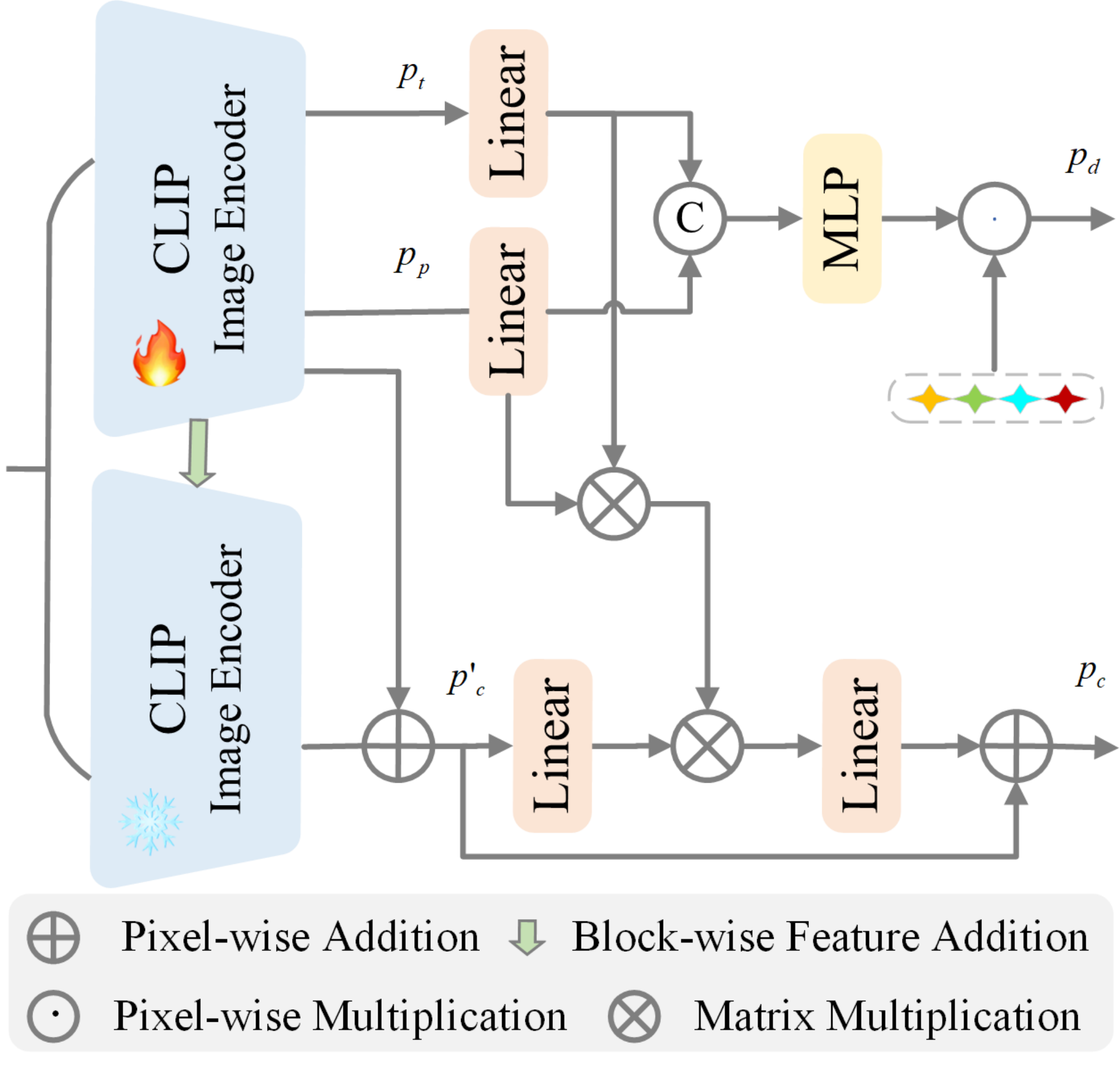}
     \caption{Illustration of our proposed latent prompt generators (LPGs). We employ the unlocked and locked image encoders of CLIP to generate the degradation prompt and caption prompt only from the processing weather-degraded images.}
     \label{fig:promptor}
 \end{figure}

As shown in Figure \ref{fig:promptor}, the output from the unlocked image encoder consists of three main components, namely the latent prompts for degradation type, degradation property, and image caption. The prompts for degradation type and property are combined and processed through multi-layer perceptron (MLP) layers followed by a scaling operation.
Therefore, we extract the degradation-specific guidance, which is then integrated into the time embedding of the diffusion model \cite{ho2020denoising}.
Furthermore, we combine the final outputs from both the unlocked and locked image encoders when generating the latent caption prompt.
These outputs are strengthened through a cross-attention mechanism that incorporates the aforementioned guidance focused on weather degradations, thereby enhancing the accuracy of caption prompt generation.

We start by training the proposed latent prompt generators with manually labeling different types and properties of weather degradations.
We utilize the bootstrapped vision-language pre-training model (BLIP) \cite{li2022blip} to generate the synthetic captions for all clean images.
In contrast to \cite{luo2023controlling}, our LPGs can effectively classify mixed weather conditions and indicate the current weather property associated with different types of weather.
Figure \ref{fig:promptor} illustrates that the image encoders take weather-degraded images as inputs, aiming to train the latent prompts generated by degraded images to closely match the degradations and image contents.
To optimize the latent prompt generators, we employ cross-entropy loss for prompt alignment. 
Our proposed latent prompt generators enables our model to focus on the targeted restoration task without relying on extra image pairs \cite{wang2023images} or prompts \cite{ma2023prores}, thereby simplifying comprehensive weather degradation removal.

\begin{figure*}[!t]
\centering
\includegraphics[width=\linewidth]{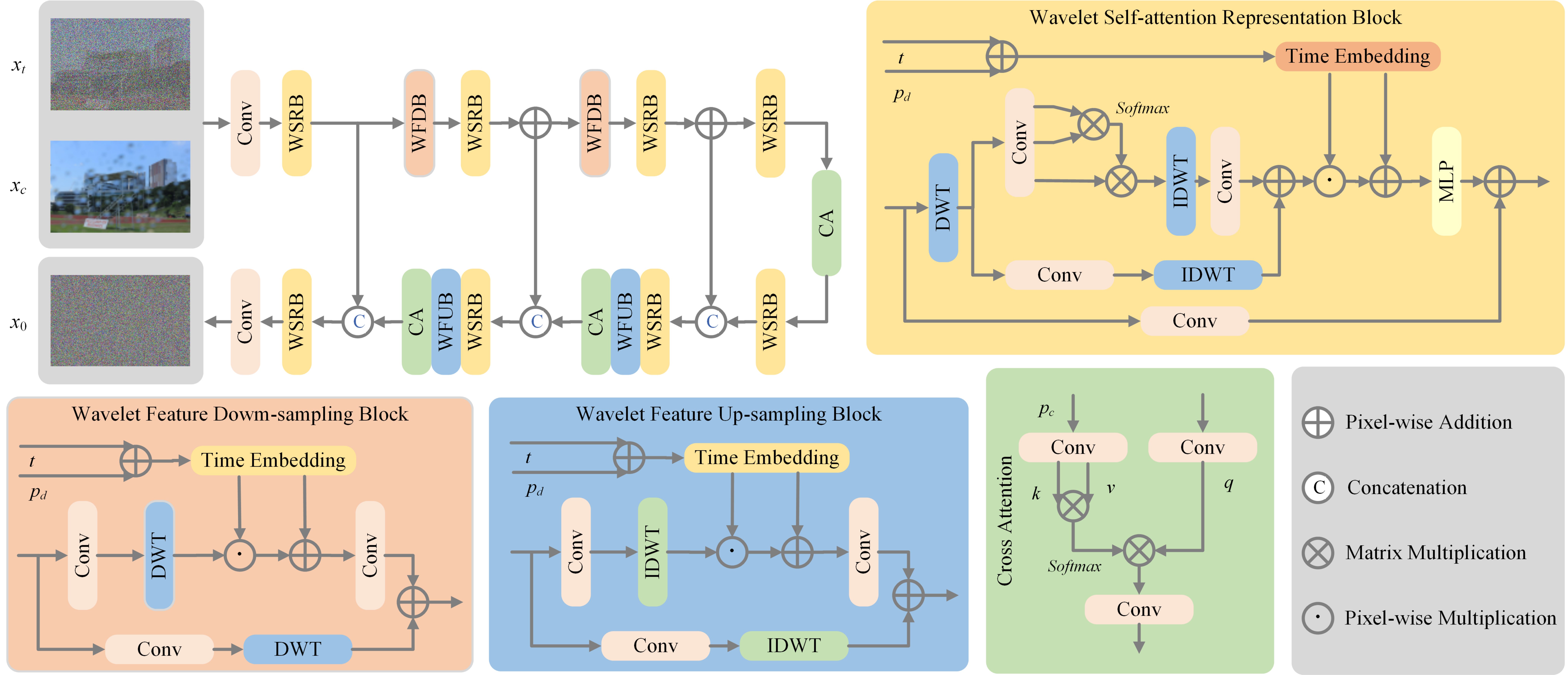}
\caption{Architectural illustration of our proposed wavelet-oriented noise estimating network (WNE-Net). We intentionally leave out the complexities of time embedding, degradation-aware prompts and caption prompts in our noise estimating network for simplification. Our WNE-Net follows a u-shaped architectural design, where the input consists of the combined state of the noisy signal $x_t$ at time $t$ and the degraded image condition $x_c$.}
\label{fig:WNE-Net}
\end{figure*}

\subsection{Adaptive Degradation-aware Self-prompting Model}

After training latent prompt generators, we utilize the image-conditioned diffusion model \cite{ho2020denoising} as the core image restoration model. As shown in Figure \ref{fig:overview}, we concatenate and process the latent prompts related to degradation type and property to create a precise degradation-aware prompt, namely
\begin{equation}
    \begin{aligned}
        & \bm{p}'_t = {\rm Linear}(\bm{p}_t), \bm{p}'_p = {\rm Linear}(\bm{p}_p), \\
        & \bm{p}_d = {\rm \alpha}({\rm MLP}({\rm Concat}(\bm{p}'_t, \bm{p}'_p)),
    \end{aligned}
\end{equation}
where $p_t$, $p_p$ and $p_d$ denotes the degradation type prompt, degradation property prompt and degradation-aware prompt, respectively.
Concat indicates the channel-wise concatenation, Linear is the linear layer, $\alpha$ is the learnable scaling parameters.
This enhanced prompt significantly improves the effectiveness of our restoration model in identifying and correcting modern degradations \cite{luo2023controlling}, making it a crucial component of our self-prompting capabilities. Additionally, we integrate the cross-attention to enhance feature representation and control the flow of contextual information, namely
\begin{equation}
    \begin{aligned}
        & \bm{A}_{prompt} = {\rm softmax}(\beta_1(\bm{p}'_t\bm{p}_p'^{\rm T})),\\
        & \bm{p}_c = {\rm Linear}(\bm{A}_{prompt}{\rm Linear}(\bm{p}'_c)) + \bm{p}'_c, \\
    \end{aligned}
\end{equation}
where softmax denotes the activation function, $\beta_1$ is a learnable scaling factor, T indicates the matrix transposing operation.
Given the consistent high-resolution nature of weather-degraded images and the exponential increase in computational demands from self-attention mechanism relative to the input feature resolution, we restrict cross-attention operations to the middle and decoder stages of noise estimating network.
The cross-attention mechanism here can be formulated as
\begin{equation}
    \begin{aligned}
        & \bm{Q}_{caption}, \bm{K}_{caption} = {\rm Conv}(\bm{p}_c), \\
        & \bm{A}_{cross} = {\rm softmax}(\beta_2(\bm{Q}_{caption}\bm{K}_{caption}^{\rm T})),\\
        & \bm{x}_{cross} = {\rm Conv}(\bm{A}_{cross}{\rm Conv}(x)), \\
    \end{aligned}
\end{equation}
where Conv denotes the $1\times 1$ convolution layer, $\beta_2$ is a learnable scaling factor.
Therefore, when encountering a noisy state $x_t$ at time $t$ and weather-degraded images $x_d$, our model incorporates prompts related to degradation type, degradation property and image caption into the noise estimating network \cite{ho2020denoising, luo2023image}, thereby the noise estimating network is modified into
\begin{equation}
    \epsilon_{\theta}(\bm{x}_{t}; \bm{x}_{d}, t) \rightarrow \epsilon_{\theta}(\bm{x}_{t}; \bm{x}_{d}, t, \bm{p}_{t}, \bm{p}_{p}, \bm{p}_{c}),
\end{equation}
where $\bm{x}_t$ is the noisy state at time $t$, $\bm{x}_d$ denotes the condition of degraded images.
Our self-prompting model makes two key contributions. Firstly, the degradation-aware prompt is designed to enhance the classification of weather degradations and specify the characteristics of current weather degradations. Additionally, the caption prompt enhances the accuracy of image reconstruction.

\subsection{Wavelet-oriented Noise Estimating Network}

As illustrated in Figure \ref{fig:WNE-Net}, our approach aims to improve the efficiency of diffusion model during inference by incorporating the wavelet transformations.

\noindent\textbf{Wavelet Self-attention Representation Block}
Traditional residual blocks commonly used in diffusion model \cite{ho2020denoising} for feature learning often overlook the integration of global information and distinct frequency components.
Inspired by \cite{mallat1989theory, phung2023wavelet}, we introduce a wavelet self-attention representation block (WSRB).
Initially, WSRB applies DWT to separate frequency components and simultaneously reduce the feature resolution. 
DWT is mathematically defined as
\begin{equation}
\begin{array}{l}
     W_{\varphi}[j+1, k]=\left.h_{\varphi}[-n] * W_{\varphi}[j, n]\right|_{n=2 k, k \leq 0},  \\
     W_{\psi}[j+1, k]=\left.h_{\psi}[-n] * W_{\psi}[j, n]\right|_{n=2 k, k \leq 0}, 
\end{array}
\end{equation}
where $\varphi$ and $\psi$ denote the approximation and detail function respectively, $W_{\varphi}$ and $W_{\psi}$ are the coefficients, $h_{\varphi}[-n]$ is the time reversed scaling vector while the $h_{\psi}[-n]$ is the wavelet vector, $n$ and $j$ denote the sampling of vector and the resolution level.
Subsequently, we employ a self-attention mechanism to enhance features and reconstruct them to their original resolution using the inverse discrete wavelet transform (IDWT), namely
\begin{equation}
    \begin{aligned}
        & \bm{Q}, \bm{K}, \bm{V} = {\rm Conv}({\rm DWT}(\bm{x})), \\
        & \bm{A}_{wsrb} = {\rm softmax}(\beta_3(\bm{Q}\bm{K}^{\rm T})), \\
        & \bm{x}_{global} = {\rm Conv}({\rm IDWT}(\bm{A}_{wsrb}\bm{V})),
    \end{aligned}
\end{equation}
where $\beta_3$ is a learnable scaling factor.
This approach effectively utilizes the diverse frequency components to create precise attention maps and reduces the computational overhead associated with self-attention.
Furthermore, we introduce two branches to merge cross-level features. One branch uses convolution to enhance individual frequency variables and integrate them with features improved by self-attention, capturing both local and global information.
Therefore, the output feature of our proposed WSRB can be obtained by
\begin{equation}
\begin{aligned}
    & \bm{x}_{local} = {\rm IDWT}({\rm Conv}({\rm DWT}(\bm{x}))), \\
    & \bm{x}' = \bm{x}_{global} + \bm{x}_{local}.
\end{aligned}
\end{equation}
This addresses the limitation of traditional self-attention mechanism in representing high-frequency dependencies \cite{gao2023frequency, bai2022improving}. Furthermore, we incorporate the skip convolution after a multi-layer perceptron to facilitate long-distance information exchange and enhance the training robustness \cite{huang2023scalelong}.
Between these branches, we integrate the degradation-aware prompt and time embedding $t$ to help the model identify degradation types and properties.
Therefore, the process of WSRB can be formulated as
\begin{equation}
    \begin{aligned}
        & t' = t + p_d, \\
        & \bm{x}'' = \bm{x}' \odot t' + t', \\
        & \bm{x}_{wsrb} = {\rm MLP}(\bm{x}'') + {\rm Conv}(\bm{x}),
    \end{aligned}
\end{equation}
where $\odot$ denotes the element-wise  multiplication, $w_{wsrb}$ is the output features of our proposed wavelet self-attention representation block.

\noindent\textbf{Wavelet Feature Sampling Block}
The conventional cross-stage sampling methods used in diffusion model often overlook frequency characteristics \cite{ho2020denoising, phung2023wavelet, luo2023image}. Meanwhile, the overlapping sampling strategies can cause misalignment, leading to the inaccurate noise estimation.
In this work, we propose a wavelet feature sampling block (WFSB) that leverages the inherent sampling properties of DWT and IDWT to improve the cross-stage progression. Unlike existing noise estimating networks, we incorporate degradation-aware prompts and time embedding during the sampling phase to better perceive the weather degradations. Additionally, we integrate the input features with enhanced features to effectively capture noise characteristics from the input and gain insights into degradations.
Specifically, our wavelet feature sampling block contains the wavelet feature down-sampling block (WFDB) and wavelet feature up-sampling block (WFUB).
The WFDB can be expressed as
\begin{equation}
    \begin{aligned}
        & \bm{x}' = {\rm DWT}({\rm Conv}(\bm{x})), \\
        & \bm{x}'' = \bm{x}'\odot t' + t', \\
        & \bm{x}_{wfdb} = {\rm Conv}(\bm{x}'') + {\rm DWT}({\rm Conv}(\bm{x})),
    \end{aligned}
\end{equation}
where $\bm{x}_{wfdb}$ is the output features of our proposed wavelet feature down-sampling block.
Evidently, the down-sampling nature of DWT reduces the feature resolution.
Meanwhile, the WFUB is derived from
\begin{equation}
    \begin{aligned}
        & \bm{x}' = {\rm IDWT}({\rm Conv}(\bm{x})), \\
        & \bm{x}'' = \bm{x}'\odot t' + t', \\
        & \bm{x}_{wfub} = {\rm Conv}(\bm{x}'') + {\rm IDWT}({\rm Conv}(\bm{x})),
    \end{aligned}
\end{equation}
where $\bm{x}_{wfub}$ is the output features of our proposed wavelet feature up-sampling block.
Similarly, the up-sampling nature of IDWT can enhance the resolution of features.

\section{Experiments}
\label{sec:experiments}

\subsection{Implementation Details}

We conduct experiments on NVIDIA Tesla A800 GPU.
The AdamW optimizer with $\beta_1=0.9$ and $\beta_2=0.99$ are utilized to optimize the parameters.
During the training process, we first train our proposed latent prompt generators, and then embed the latent prompts generated by the latent prompt generators into wavelet-aware noise estimating network to conduct further training.
We set the number of learning iterations to 10 000 on latent prompt generators, and the number of prompted iterations of image reconstruction to 1 000 000.
The learning rates of latent prompt generators and diffusion-based restoration model are set to $5\times 10^{-6}$ and $2\times 10^{-4}$, respectively.
The latter is gradually decreased to $1\times 10^{-6}$ by utilizing cosine annealing.

\begin{table}[h]\footnotesize
    \centering
    \caption{Partitioning details of the training and testing samples in the collected AWIR120K dataset with different weather degradations.}
    \renewcommand\arraystretch{1.15}
    \setlength{\tabcolsep}{1.5mm}{
    \begin{tabular}{llcc}
    \toprule
    Type & Dataset & Training Samples & Testing Samples \\
    \midrule
    Rain & Rain200H & 1 800 & 200 \\
    Haze & RESIDE-6K & 6 000 & 1 000 \\
    Raindrop & Raindrop-A & 861 & 58 \\
    Snow & Snow100K-L & 50 000 & 16 801 \\
    low-lightness & LOL & 485 & 15 \\
    Rain-by-snow & RS100K-L & 50 000 & 2 000 \\
    Rain-by-haze & Test1 & 9 000 & 750 \\
    Rain-by-raindrop & RainDS-syn & 1 000 & 200 \\
    \bottomrule
    \end{tabular}}
    \label{tab:AWIR120K}
\end{table}

\subsection{Datasets and Evaluation Metrics}
We experimentally evaluate our approach on the images degraded by single and mingled weather degradations, including rain (Rain200H \cite{yang2017deep}), haze (RESIDE-6K \cite{li2018benchmarking}), raindrop (Raindrop-A \cite{qian2018attentive}), snow (Snow100K-L \cite{liu2018desnownet}), low-lightness (LOL \cite{wei2018deep}), rain-by-snow (RS100K-L \cite{wen2024restoring}), rain-by-haze (Test1 \cite{li2019heavy}) and rain-by-raindrop (RainDS-syn \cite{quan2021removing}). To train our proposed model, we compose the comprehensive AWIR120K dataset, which contains 119 146 image pairs and eight weather degradations.
The specific setting of AWIR120K dataset is shown in Table \ref{tab:AWIR120K}.

We calculate the peak signal-to-noise ratio (PSNR), structural similarity (SSIM) and learned perpetual image patch similarity (LPIPS) \cite{zhang2018unreasonable} between the reconstructed images and corresponding ground truth, where the higher metrical scores indicates to better image quality.
Meanwhile, we employ the naturalness image quality evaluator (NIQE) \cite{mittal2012making} to quantitatively evaluate the naturalness, where the better perceptual image quality corresponds to lower indicator.

\begin{table*}[!t]\footnotesize
\centering
\caption{Quantitative comparisons of the task-specific methods. The \red{\textbf{red}} and \blue{\textbf{blue}} metrical scores denote the best and second-best quantitative performance. Our proposed ADSM achieves the best metrical scores across the eight testing datasets.}
\label{tab:specific}
\begin{minipage}{0.23\linewidth}
    \centering
    \caption*{(a) Image deraining}
    \renewcommand\arraystretch{1.15}
    \setlength{\tabcolsep}{1.7mm}{
    \begin{tabular}{lcc}
    \toprule
    \multirow{2}{*}{Method} & \multicolumn{2}{c}{Rain200H} \\
    ~ & PSNR & SSIM  \\
    \midrule
    RESCAN \cite{li2018recurrent} & 26.75 & 0.835 \\
    PReNet \cite{ren2019progressive} & 29.04 & 0.899 \\
    MSPFN \cite{jiang2020multi} & 29.36 & 0.903 \\
    MPRNet \cite{zamir2021multi} & 30.67 & 0.911 \\
    Uformer \cite{wang2022uformer} & 30.80 & 0.911 \\
    IDT \cite{xiao2022image} & 32.10 & \blue{\textbf{0.934}} \\
    DRSformer \cite{chen2023learning} & \blue{\textbf{32.17}} & 0.933 \\
    \textbf{ADSM (ours)} & \red{\textbf{32.23}} & \red{\textbf{0.937}} \\
    \bottomrule
    \end{tabular}
    }
\end{minipage}
\hfill
\begin{minipage}{0.23\linewidth}
    \centering
    \caption*{(b) Image dehazing}
    \renewcommand\arraystretch{1.15}
    \setlength{\tabcolsep}{0.8mm}{
    \begin{tabular}{lcc}
    \toprule
    \multirow{2}{*}{Method} & \multicolumn{2}{c}{RESIDE-6K} \\
    ~ & PSNR & SSIM \\
    \midrule
    AOD-Net \cite{li2017aod} & 20.27 & 0.855 \\
    GCANet \cite{das2022gca} & 25.09 & 0.923 \\
    GridDehazeNet \cite{liu2019griddehazenet} & 25.86 & 0.944 \\
    FFA-Net \cite{qin2020ffa} & 29.96 & \blue{\textbf{0.973}} \\
    MSBDN \cite{dong2020multi} & 28.56 & 0.966 \\
    AECR-Net \cite{wu2021contrastive} & 28.52 & 0.964 \\
    DehazeFormer \cite{song2023vision} & \blue{\textbf{30.29}} & 0.964 \\
    \textbf{ADSM (ours)} & \red{\textbf{30.66}} & \red{\textbf{0.980}} \\
    \bottomrule
    \end{tabular}
    }
\end{minipage}
\hfill
\begin{minipage}{0.23\linewidth}
    \centering
    \caption*{(c) Raindrop removal}
    \renewcommand\arraystretch{1.15}
    \setlength{\tabcolsep}{1mm}{
    \begin{tabular}{lcc}
    \toprule
    \multirow{2}{*}{Method} & \multicolumn{2}{c}{Raindrop-A} \\
    ~ & PSNR & SSIM \\
    \midrule
    RadNet \cite{wei2022robust} & 24.54 & 0.885 \\
    attentiveGAN \cite{qian2018attentive} & 31.57 & 0.902 \\
    DuRN \cite{liu2019dual} & 31.24 & 0.926 \\
    SwinIR \cite{liang2021swinir} & 30.82 & 0.904 \\
    MPRNet \cite{zamir2021multi} & 32.15 & 0.937 \\
    IDT \cite{xiao2022image} & 31.87 & 0.931 \\
    WeatherDiff \cite{ozdenizci2023restoring} & \blue{\textbf{32.29}} & \blue{\textbf{0.942}} \\
    \textbf{ADSM (ours)} & \red{\textbf{33.42}} & \red{\textbf{0.947}} \\
    \bottomrule
    \end{tabular}
    }
\end{minipage}
\hfill
\begin{minipage}{0.23\linewidth}
    \centering
    \caption*{(d) Image desnowing}
    \renewcommand\arraystretch{1.15}
    \setlength{\tabcolsep}{1mm}{
    \begin{tabular}{lcc}
    \toprule
    \multirow{2}{*}{Method} & \multicolumn{2}{c}{Snow100K-L} \\
    ~ & PSNR & SSIM \\
    \midrule
    RESCAN \cite{li2018recurrent} & 26.08 & 0.811 \\
    DesnowNet \cite{liu2018desnownet} & 27.17 & 0.898 \\
    SPANet \cite{wang2019spatial} & 23.70 & 0.793 \\
    JSTASR \cite{chen2020jstasr} & 25.32 & 0.808 \\
    DDMSNet \cite{zhang2021deep} & 28.85 & 0.877 \\
    DesnowGAN \cite{jaw2020desnowgan} & 28.07 & \blue{\textbf{0.921}} \\
    WeatherDiff \cite{ozdenizci2023restoring} & \blue{\textbf{30.28}} & 0.900 \\
    \textbf{ADSM (ours)} & \red{\textbf{31.14}} & \red{\textbf{0.939}} \\
    \bottomrule
    \end{tabular}
    }
\end{minipage}
\hfill
\begin{minipage}{0.23\linewidth}
    \centering
    \vspace{3mm}
    \caption*{(e) Image enhancement}
    \renewcommand\arraystretch{1.15}
    \setlength{\tabcolsep}{0.8mm}{
    \begin{tabular}{lcc}
    \toprule
    \multirow{2}{*}{Method} & \multicolumn{2}{c}{LOL} \\
    ~ & PSNR & SSIM \\
    \midrule
    RetinexNet \cite{wei2018deep} & 17.56 & 0.698 \\
    EnlightenGAN \cite{jiang2021enlightengan} & 17.61 & 0.653 \\
    RUAS \cite{liu2021retinex} & 16.40 & 0.503 \\
    MIRNet \cite{zamir2020learning} & \blue{\textbf{24.14}} & \blue{\textbf{0.830}} \\
    URetinex-Net \cite{wu2022uretinex} & 19.84 & 0.824 \\
    IR-SDE \cite{luo2023image} & 20.45 & 0.787 \\
    LLFormer \cite{wang2023ultra} & 23.65 & 0.816 \\
    \textbf{ADSM (ours)} & \red{\textbf{24.92}} & \red{\textbf{0.885}} \\
    \bottomrule
    \end{tabular}
    }
\end{minipage}
\hfill
\begin{minipage}{0.23\linewidth}
    \centering
    \vspace{3mm}
    \caption*{(f) Rain-by-snow removal}
    \renewcommand\arraystretch{1.15}
    \setlength{\tabcolsep}{1.3mm}{
    \begin{tabular}{lcc}
    \toprule
    \multirow{2}{*}{Method} & \multicolumn{2}{c}{RS100K-L} \\
    ~ & PSNR & SSIM \\
    \midrule
    JSTASR \cite{chen2020jstasr} & 25.74 & 0.862 \\
    DesnowNet \cite{liu2018desnownet} & 27.31 & 0.913 \\
    PReNet \cite{ren2019progressive} & 26.56 & 0.907 \\
    TransWeather \cite{valanarasu2022transweather} & 28.97 & 0.921 \\
    MPRNet \cite{zamir2021multi} & \blue{\textbf{29.59}} & 0.931 \\
    Uformer \cite{wang2022uformer} & 25.49 & 0.890 \\
    SnowFormer \cite{chen2022snowformer} & 29.10 & \blue{\textbf{0.932}} \\
    \textbf{ADSM (ours)} & \red{\textbf{30.71}} & \red{\textbf{0.943}} \\
    \bottomrule
    \end{tabular}
    }
\end{minipage}
\hfill
\begin{minipage}{0.23\linewidth}
    \centering
    \vspace{3mm}
    \caption*{(g) Rain-by-haze removal}
    \renewcommand\arraystretch{1.15}
    \setlength{\tabcolsep}{1.3mm}{
    \begin{tabular}{lcc}
    \toprule
    \multirow{2}{*}{Method} & \multicolumn{2}{c}{Test1} \\
    ~ & PSNR & SSIM \\
    \midrule
    CycleGAN \cite{zhu2017unpaired} & 17.62 & 0.656 \\
    pix2pix \cite{isola2017image} & 19.09 & 0.710 \\
    HRGAN \cite{li2019heavy} &  21.56 & 0.855 \\
    SwinIR \cite{liang2021swinir} & 23.23 & 0.869 \\
    PCNet \cite{jiang2021rain} & 26.19 & 0.902 \\
    MPRNet \cite{zamir2021multi} &  28.03 & 0.919 \\
    WeatherDiff \cite{ozdenizci2023restoring} &  \blue{\textbf{28.38}} & \blue{\textbf{0.932}} \\
    \textbf{ADSM (ours)} & \red{\textbf{30.26}} & \red{\textbf{0.937}} \\
    \bottomrule
    \end{tabular}
    }
\end{minipage}
\hfill
\begin{minipage}{0.23\linewidth}
    \centering
    \vspace{3mm}
    \caption*{(h) Rain-by-raindrop removal}
    \renewcommand\arraystretch{1.15}
    \setlength{\tabcolsep}{1.7mm}{
    \begin{tabular}{lcc}
    \toprule
    \multirow{2}{*}{Method} & \multicolumn{2}{c}{RainDS-syn} \\
    ~ & PSNR & SSIM \\
    \midrule
    RESCAN \cite{li2018recurrent} & 27.43 & 0.818 \\
    MPRNet \cite{zamir2021multi} & 34.99 & 0.956 \\
    CCN \cite{quan2021removing} & 34.79 & 0.957 \\
    DGUNet \cite{mou2022deep} & 35.34 & 0.959 \\
    Uformer \cite{wang2022uformer} & 34.99 & 0.954 \\
    NAFNet \cite{chen2022simple} & 34.99 & 0.957 \\
    IDT \cite{xiao2022image} & \blue{\textbf{36.23}} & \blue{\textbf{0.960}} \\
    \textbf{ADSM (ours)} & \red{\textbf{36.88}} & \red{\textbf{0.972}} \\
    \bottomrule
    \end{tabular}
    }
\end{minipage}
\end{table*}

\begin{table*}[htp]\footnotesize
\centering
\caption{Quantitative comparisons of the all-in-one methods. Our proposed ADSM achieves the best metrical scores across the eight weather-degraded datasets.}
\label{tab:all-in-one}
\begin{minipage}{0.325\linewidth}
    \centering
    \caption*{(a) Image deraining}
    \renewcommand\arraystretch{1.15}
    \setlength{\tabcolsep}{0.7mm}{
    \begin{tabular}{lcccc}
    \toprule
    \multirow{2}{*}{Method} & \multicolumn{4}{c}{Rain200H} \\
    ~ & PSNR & SSIM & LPIPS & NIQE  \\
    \midrule
    Restormer \cite{zamir2022restormer} & 28.11 & 0.8845 & 0.1334 & 5.032 \\
    DehazeFormer \cite{song2023vision} & 27.37 & 0.8477 & 0.1758 & 4.550 \\
    SnowFormer \cite{chen2022snowformer} &  26.42 & 0.8467 & 0.1554 & 4.348 \\
    LLFormer \cite{wang2023ultra} & 26.62 & 0.8310 & 0.2123 & 4.469 \\
    IDT \cite{xiao2022image} & 28.13 & 0.8593 & 0.1605 & 4.633 \\
    SASCFormer \cite{wen2024restoring} & 26.76 & 0.8413 & 0.1774 & 4.081 \\
    UDSR$^2$Former \cite{chen2023sparse} & 27.67 & 0.8562 & 0.1566 & 4.750 \\
    \midrule
    All-in-One \cite{li2020all} & 21.55 & 0.7440 & 0.3219 & 4.310 \\
    TKL \cite{chen2022learning} & 26.14 & 0.8463 & 0.2205 & 5.767 \\
    TransWeather \cite{valanarasu2022transweather} & 25.74 & 0.8302 & 0.1653 & 4.881 \\
    AIRFormer \cite{gao2023frequency} & 24.46 & 0.7424 & 0.3154 & 7.129 \\
    WeatherDiff \cite{ozdenizci2023restoring} & 26.66 & 0.8541 & \blue{\textbf{0.1071}} & \blue{\textbf{4.007}} \\
    PromptIR \cite{potlapalli2023promptir} & \blue{\textbf{29.29}} & \blue{\textbf{0.8890}} & 0.1176 & 4.806 \\
    \textbf{ADSM (ours)} & \red{\textbf{29.74}} & \red{\textbf{0.8938}} & \red{\textbf{0.0934}} & \red{\textbf{3.917}} \\
    \bottomrule
    \end{tabular}
    }
\end{minipage}
\hfill
\begin{minipage}{0.325\linewidth}
    \centering
    \caption*{(b) Image dehazing}
    \renewcommand\arraystretch{1.15}
    \setlength{\tabcolsep}{0.7mm}{
    \begin{tabular}{lcccc}
    \toprule
    \multirow{2}{*}{Method} & \multicolumn{4}{c}{RESIDE-6K} \\
    ~ & PSNR & SSIM & LPIPS & NIQE \\
    \midrule
    Restormer \cite{zamir2022restormer} & 28.96 & 0.9774 & 0.0344 & 4.733 \\
    DehazeFormer \cite{song2023vision} & 28.74 & 0.9744 & 0.0333 & 4.546 \\
    SnowFormer \cite{chen2022snowformer} & 28.45 & 0.9739 & 0.0352 & 4.595 \\
    LLFormer \cite{wang2023ultra} & 26.94 & 0.9649 & 0.0412 & 4.492 \\
    IDT \cite{xiao2022image} & 28.93 & 0.9735 & 0.0373 & 4.450 \\
    SASCFormer \cite{wen2024restoring} & 22.73 & 0.9015 & 0.1273 & \blue{\textbf{4.176}} \\
    UDSR$^2$Former \cite{chen2023sparse} & 26.45 & 0.9652 & 0.0479 & 4.407 \\
    \midrule
    All-in-One \cite{li2020all} & 19.31 & 0.8761 & 0.1689 & 4.559 \\
    TKL \cite{chen2022learning} & 25.52 & 0.9619 & 0.0497 & 4.553 \\
    TransWeather \cite{valanarasu2022transweather} &  26.69 & 0.9653 & 0.0368 & 4.406 \\
    AIRFormer \cite{gao2023frequency} & 25.67 & 0.9513 & 0.0639 & 4.302 \\
    WeatherDiff \cite{ozdenizci2023restoring} & 24.30 & 0.9540 & 0.0514 & 4.541 \\
    PromptIR \cite{potlapalli2023promptir} & \blue{\textbf{29.68}} & \blue{\textbf{0.9780}} & \blue{\textbf{0.0278}} & 4.527 \\
    \textbf{ADSM (ours)} & \red{\textbf{30.66}} & \red{\textbf{0.9870}} & \red{\textbf{0.0202}} & \red{\textbf{4.103}} \\
    \bottomrule
    \end{tabular}
    }
\end{minipage}
\hfill
\begin{minipage}{0.325\linewidth}
    \centering
    \caption*{(c) Raindrop removal}
    \renewcommand\arraystretch{1.15}
    \setlength{\tabcolsep}{0.7mm}{
    \begin{tabular}{lcccc}
    \toprule
    \multirow{2}{*}{Method} & \multicolumn{4}{c}{Raindrop-A} \\
    ~ & PSNR & SSIM & LPIPS & NIQE \\
    \midrule
    Restormer \cite{zamir2022restormer} & 30.84 & \blue{\textbf{0.9340}} & \blue{\textbf{0.0874}} & 4.086 \\
    DehazeFormer \cite{song2023vision} & 30.09 & 0.9187 & 0.1108 & 3.898 \\
    SnowFormer \cite{chen2022snowformer} & 30.07 & 0.9226 & 0.1040 & 4.193 \\
    LLFormer \cite{wang2023ultra} & 30.12 & 0.9149 & 0.1257 & 4.032 \\
    IDT \cite{xiao2022image} & 29.72 & 0.9216 & 0.1014 & 3.836 \\
    SASCFormer \cite{wen2024restoring} & 28.99 & 0.9081 & 0.1274 & 3.640 \\
    UDSR$^2$Former \cite{chen2023sparse} & 29.61 & 0.9209 & 0.1145 & 4.197 \\
    \midrule
    All-in-One \cite{li2020all} & 23.41 & 0.8722 & 0.2099 & 4.099 \\
    TKL \cite{chen2022learning} & 27.39 & 0.9170 & 0.1085 & 3.773 \\
    TransWeather \cite{valanarasu2022transweather} & 28.83 & 0.9118 & 0.0891 & 4.078 \\
    AIRFormer \cite{gao2023frequency} & 27.21 & 0.8777 & 0.1808 & \blue{\textbf{3.619}} \\
    WeatherDiff \cite{ozdenizci2023restoring} & 25.08 & 0.8934 & 0.1096 & 3.818 \\
    PromptIR \cite{potlapalli2023promptir} & \blue{\textbf{31.31}} & 0.9320 & 0.0877 & 3.943 \\
    \textbf{ADSM (ours)} & \red{\textbf{31.94}} & \red{\textbf{0.9385}} & \red{\textbf{0.0699}} & \red{\textbf{3.524}} \\
    \bottomrule
    \end{tabular}
    }
\end{minipage}
\hfill
\begin{minipage}{0.325\linewidth}
    \centering
    \vspace{3mm}
    \caption*{(d) Image desnowing}
    \renewcommand\arraystretch{1.15}
    \setlength{\tabcolsep}{0.7mm}{
    \begin{tabular}{lcccc}
    \toprule
    \multirow{2}{*}{Method} & \multicolumn{4}{c}{Snow100K-L} \\
    ~ & PSNR & SSIM & LPIPS & NIQE \\
    \midrule
    Restormer \cite{zamir2022restormer} & 31.51 & 0.9270 & 0.0858 & 4.393 \\
    DehazeFormer \cite{song2023vision} & 30.86 & 0.9107 & 0.1060 & 4.107 \\
    SnowFormer \cite{chen2022snowformer} & 29.86 & 0.9122 & 0.1020 & 4.031 \\
    LLFormer \cite{wang2023ultra} & 30.74 & 0.9099 & 0.1053 & 3.949 \\
    IDT \cite{xiao2022image} & 31.68 & 0.9187 & 0.0907 & 4.017 \\
    SASCFormer \cite{wen2024restoring} & 30.34 & 0.9081 & 0.1067 & 3.759 \\
    UDSR$^2$Former \cite{chen2023sparse} & 31.73 & 0.9215 & 0.0923 & 4.204 \\
    \midrule
    All-in-One \cite{li2020all} & 26.69 & 0.8829 & 0.1663 & 3.946 \\
    TKL \cite{chen2022learning} & 30.55 & 0.9176 & 0.0989 & 4.278 \\
    TransWeather \cite{valanarasu2022transweather} & 30.10 & 0.9039 & 0.0907 & 3.854 \\
    AIRFormer \cite{gao2023frequency} & 28.56 & 0.8684 & 0.1678 & 3.853 \\
    WeatherDiff \cite{ozdenizci2023restoring} &  28.99 & 0.8941 & 0.0880 & \blue{\textbf{3.733}} \\
    PromptIR \cite{potlapalli2023promptir} & \blue{\textbf{32.14}} & \blue{\textbf{0.9282}} & \blue{\textbf{0.0841}} & 4.392 \\
    \textbf{ADSM (ours)} & \red{\textbf{32.85}} & \red{\textbf{0.9291}} & \red{\textbf{0.0718}} & \red{\textbf{3.700}} \\
    \bottomrule
    \end{tabular}
    }
\end{minipage}
\hfill
\begin{minipage}{0.325\linewidth}
    \centering
    \vspace{3mm}
    \caption*{(e) Image enhancement}
    \renewcommand\arraystretch{1.15}
    \setlength{\tabcolsep}{0.7mm}{
    \begin{tabular}{lcccc}
    \toprule
    \multirow{2}{*}{Method} & \multicolumn{4}{c}{LOL} \\
    ~ & PSNR & SSIM & LPIPS & NIQE \\
    \midrule
    Restormer \cite{zamir2022restormer} & 21.07 & 0.8940 & 0.1700 & 4.659 \\
    DehazeFormer \cite{song2023vision} & \blue{\textbf{24.14}} & 0.8907 & 0.1955 & 4.976 \\
    SnowFormer \cite{chen2022snowformer} &  21.62 & 0.8697 & 0.1915 & 4.673 \\
    LLFormer \cite{wang2023ultra} & 24.00 & 0.8794 & 0.2036 & 4.206 \\
    IDT \cite{xiao2022image} & 21.18 & 0.8778 & 0.1919 & \blue{\textbf{3.745}} \\
    SASCFormer \cite{wen2024restoring} & 22.54 & 0.8744 & 0.2047 & 3.880 \\
    UDSR$^2$Former \cite{chen2023sparse} & 20.39 & 0.8647 & 0.2287 & 4.316 \\
    \midrule
    All-in-One \cite{li2020all} & 18.95 & 0.8213 & 0.4023 & 4.333 \\
    TKL \cite{chen2022learning} & 20.58 & 0.8734 & 0.2235 & 4.630 \\
    TransWeather \cite{valanarasu2022transweather} & 23.46 & 0.8896 & 0.1677 & 4.645 \\
    AIRFormer \cite{gao2023frequency} & 22.23 & 0.8312 & 0.3195 & 4.146 \\
    WeatherDiff \cite{ozdenizci2023restoring} & 18.13 & 0.8426 & 0.2262 & 4.962 \\
    PromptIR \cite{potlapalli2023promptir} & 24.08 & \blue{\textbf{0.8999}} & \blue{\textbf{0.1415}} & 4.268 \\
    \textbf{ADSM (ours)} & \red{\textbf{25.14}} & \red{\textbf{0.9022}} & \red{\textbf{0.1399}} & \red{\textbf{3.583}} \\
    \bottomrule
    \end{tabular}
    }
\end{minipage}
\hfill
\begin{minipage}{0.325\linewidth}
    \centering
    \vspace{3mm}
    \caption*{(f) Rain-by-snow removal}
    \renewcommand\arraystretch{1.15}
    \setlength{\tabcolsep}{0.7mm}{
    \begin{tabular}{lcccc}
    \toprule
    \multirow{2}{*}{Method} & \multicolumn{4}{c}{RS100K-L} \\
    ~ & PSNR & SSIM & LPIPS & NIQE \\
    \midrule
    Restormer \cite{zamir2022restormer} & 30.98 & 0.9205 & 0.0942 & 4.465 \\
    DehazeFormer \cite{song2023vision} & 30.35 & 0.9022 & 0.1172 & 4.224 \\
    SnowFormer \cite{chen2022snowformer} & 29.23 & 0.9032 & 0.1110 & 4.123 \\
    LLFormer \cite{wang2023ultra} & 30.13 & 0.8998 & 0.1187 & 4.060 \\
    IDT \cite{xiao2022image} & 31.13 & 0.9103 & 0.1017 & 4.121 \\
    SASCFormer \cite{wen2024restoring} & 29.95 & 0.8998 & 0.1183 & \blue{\textbf{3.885}} \\
    UDSR$^2$Former \cite{chen2023sparse} & 31.21 & 0.9131 & 0.1049 & 4.317 \\
    \midrule
    All-in-One \cite{li2020all} & 26.35 & 0.8711 & 0.1825 & 4.096 \\
    TKL \cite{chen2022learning} & 30.00 & 0.9081 & 0.1136 & 4.396 \\
    TransWeather \cite{valanarasu2022transweather} & 29.37 & 0.8925 & 0.1026 & 3.988 \\
    AIRFormer \cite{gao2023frequency} & 28.01 & 0.8540 & 0.1898 & 4.127 \\
    WeatherDiff \cite{ozdenizci2023restoring} & 28.86 & 0.8891 & 0.1014 & 3.993 \\
    PromptIR \cite{potlapalli2023promptir} & \blue{\textbf{31.71}} & \blue{\textbf{0.9230}} & \blue{\textbf{0.0909}} & 4.474 \\
    \textbf{ADSM (ours)} & \red{\textbf{32.24}} & \red{\textbf{0.9271}} & \red{\textbf{0.0897}} & \red{\textbf{3.765}} \\
    \bottomrule
    \end{tabular}
    }
\end{minipage}
\hfill
\begin{minipage}{0.325\linewidth}
    \centering
    \vspace{3mm}
    \caption*{(g) Rain-by-haze removal}
    \renewcommand\arraystretch{1.15}
    \setlength{\tabcolsep}{0.7mm}{
    \begin{tabular}{lcccc}
    \toprule
    \multirow{2}{*}{Method} & \multicolumn{4}{c}{Test1} \\
    ~ & PSNR & SSIM & LPIPS & NIQE \\
    \midrule
    Restormer \cite{zamir2022restormer} & 28.01 & 0.9166 & 0.1061 & 4.629 \\
    DehazeFormer \cite{song2023vision} & 29.03 & 0.8919 & 0.1302 & 4.160 \\
    SnowFormer \cite{chen2022snowformer} & 27.81 & 0.8888 & 0.1185 & 4.343 \\
    LLFormer \cite{wang2023ultra} & 27.72 & 0.8778 & 0.1557 & 4.146 \\
    IDT \cite{xiao2022image} & 29.80 & 0.9016 & 0.1184 & 4.328 \\
    SASCFormer \cite{wen2024restoring} & 26.67 & 0.8770 & 0.1394 & 3.928 \\
    UDSR$^2$Former \cite{chen2023sparse} & 28.86 & 0.9001 & 0.1160 & 4.158 \\
    \midrule
    All-in-One \cite{li2020all} & 19.33 & 0.8012 & 0.3171 & 4.436 \\
    TKL \cite{chen2022learning} & 25.76 & 0.8852 & 0.1775 & 4.600 \\
    TransWeather \cite{valanarasu2022transweather} & 26.91 & 0.8682 & 0.1196 & 3.941 \\
    AIRFormer \cite{gao2023frequency} & 24.52 & 0.7837 & 0.3025 & 4.491 \\
    WeatherDiff \cite{ozdenizci2023restoring} & 25.78 & 0.8990 & \blue{\textbf{0.0736}} & \blue{\textbf{3.788}} \\
    PromptIR \cite{potlapalli2023promptir} & \blue{\textbf{30.00}} & \blue{\textbf{0.9168}} & 0.0900 & 4.413 \\
    \textbf{ADSM (ours)} & \red{\textbf{30.72}} & \red{\textbf{0.9215}} & \red{\textbf{0.0680}} & \red{\textbf{3.714}} \\
    \bottomrule
    \end{tabular}
    }
\end{minipage}
\hfill
\begin{minipage}{0.325\linewidth}
    \centering
    \vspace{3mm}
    \caption*{(h) Rain-by-raindrop removal}
    \renewcommand\arraystretch{1.15}
    \setlength{\tabcolsep}{0.7mm}{
    \begin{tabular}{lcccc}
    \toprule
    \multirow{2}{*}{Method} & \multicolumn{4}{c}{RainDS-syn} \\
    ~ & PSNR & SSIM & LPIPS & NIQE \\
    \midrule
    Restormer \cite{zamir2022restormer} & 29.48 & 0.9147 & 0.0983 & 4.559 \\
    DehazeFormer \cite{song2023vision} & 28.63 & 0.8693 & 0.1653 & 4.678 \\
    SnowFormer \cite{chen2022snowformer} & 27.70 & 0.8673 & 0.1683 & 4.518 \\
    LLFormer \cite{wang2023ultra} & 27.44 & 0.8477 & 0.2016 & 4.723 \\
    IDT \cite{xiao2022image} & 29.30 & 0.8855 & 0.1408 & 4.399 \\
    SASCFormer \cite{wen2024restoring} & 29.06 & 0.8779 & 0.1514 & \blue{\textbf{4.074}} \\
    UDSR$^2$Former \cite{chen2023sparse} & 28.63 & 0.8822 & 0.1464 & 4.642 \\
    \midrule
    All-in-One \cite{li2020all} & 20.81 & 0.6950 & 0.3980 & 5.603 \\
    TKL \cite{chen2022learning} & 26.69 & 0.8642 & 0.1820 & 4.856 \\
    TransWeather \cite{valanarasu2022transweather} & 27.03 & 0.8428 & 0.1453 & 5.185 \\
    AIRFormer \cite{gao2023frequency} & 24.25 & 0.7174 & 0.3467 & 6.511 \\
    WeatherDiff \cite{ozdenizci2023restoring} & 27.64 & 0.8981 & \blue{\textbf{0.0927}} & 4.257 \\
    PromptIR \cite{potlapalli2023promptir} & \blue{\textbf{30.49}} & \blue{\textbf{0.9189}} & 0.0974 & 4.422 \\
    \textbf{ADSM (ours)} & \red{\textbf{30.86}} & \red{\textbf{0.9262}} & \red{\textbf{0.0828}} & \red{\textbf{3.934}} \\
    \bottomrule
    \end{tabular}
    }
\end{minipage}
\hfill
\begin{minipage}{0.325\linewidth}
    \centering
    \vspace{3mm}
    \caption*{(i) Adverse weather removal}
    \renewcommand\arraystretch{1.15}
    \setlength{\tabcolsep}{0.7mm}{
    \begin{tabular}{lcccc}
    \toprule
    \multirow{2}{*}{Method} & \multicolumn{4}{c}{Average} \\
    ~ & PSNR & SSIM & LPIPS & NIQE \\
    \midrule
    Restormer \cite{zamir2022restormer} & 28.62 & 0.9211 & 0.1012 & 4.570 \\
    DehazeFormer \cite{song2023vision} & 28.65 & 0.9007 & 0.1293 & 4.392 \\
    SnowFormer \cite{chen2022snowformer} & 27.65 & 0.8981 & 0.1232 & 4.353  \\
    LLFormer \cite{wang2023ultra} & 27.96 & 0.8907 & 0.1455 & 4.260 \\
    IDT \cite{xiao2022image} & 28.73 & 0.9060 & 0.1178 & 4.191 \\
    SASCFormer \cite{wen2024restoring} & 27.13 & 0.8860 & 0.1441 & \blue{\textbf{3.928}} \\
    UDSR$^2$Former \cite{chen2023sparse} & 28.07 & 0.9030 & 0.1259 & 4.374 \\
    \midrule
    All-in-One \cite{li2020all} & 22.05 & 0.8205 & 0.2709 & 4.423 \\
    TKL \cite{chen2022learning} & 26.58 & 0.8967 & 0.1468 & 4.607 \\
    TransWeather \cite{valanarasu2022transweather} & 27.27 & 0.8880 & 0.1146 & 4.372 \\
    AIRFormer \cite{gao2023frequency} & 25.61 & 0.8283 & 0.2358 & 4.772 \\
    WeatherDiff \cite{ozdenizci2023restoring} & 25.68 & 0.8906 & 0.1063 & 4.137 \\
    PromptIR \cite{potlapalli2023promptir} & \blue{\textbf{30.17}} & \blue{\textbf{0.9222}} & \blue{\textbf{0.0919}} & 4.479 \\
    \textbf{ADSM (ours)} & \red{\textbf{30.52}} & \red{\textbf{0.9282}} & \red{\textbf{0.0795}} & \red{\textbf{3.780}} \\
    \bottomrule
    \end{tabular}
    }
\end{minipage}
\end{table*}

\subsection{Comparisons with Task-specific Setting}
To assess our ADSM as task-specific method for weather-degraded image restoration, we conduct separate training and testing across eight weather-degraded datasets. 

\noindent\textbf{Image Deraining} Table \ref{tab:specific}(a) indicates the quantitative comparisons of the task-specific image deraining methods. As illustrated, our proposed ADSM advances the recent state-of-the-art DRSformer \cite{chen2023learning} by 0.06 dB in PSNR. Meanwhile, our approach also achieves the best SSIM score.

\noindent\textbf{Image Dehazing} As Table \ref{tab:specific}(b) reported, our model obtains the best quantitative performance on image dehazing task. Specifically, our proposed ADSM shows a performance gain of 0.37 dB over DehazeFormer \cite{song2023vision} on RESIDE-6K dataset.

\noindent\textbf{Raindrop Removal} We report the quantitative comparisons of the involved evaluated methods for raindrop removal in Table \ref{tab:specific}(c). Our approach obtains a performance improvement by 1.13 dB over WeatherDiff \cite{ozdenizci2023restoring} on Raindrop-A dataset.

\noindent\textbf{Image Desnowing} For image desnowing, our proposed ADSM achieves a significant improvement in both PSNR and SSIM scores. Specifically, our model advances recent WeatherDiff \cite{ozdenizci2023restoring} by 0.86 dB in PSNR on Snow100K-L dataset.

\noindent\textbf{Image Enhancement} On LOL dataset \cite{wei2018deep}, our method achieves the best quantitative performance. As Table \ref{tab:specific}(e) illustrated, MIRNet \cite{zamir2020learning} obtains the second-best performance in PSNR, and our method achieves a performance improvement by 0.78 dB.

\noindent\textbf{Rain-by-snow Removal} To evaluate our model in rain-by-snow weather removal, we conduct comparative experiments of the several methods and our proposed method. As Table \ref{tab:specific}(f) shown, our model obtains performance gains of 0.71 dB over MPRNet \cite{zamir2021multi} in PSNR indicator. 

\noindent\textbf{Rain-by-haze Removal} In Table \ref{tab:specific}(g), we compare our proposed ADSM and the other involved methods. As reported, the metrical scores obtained by our approach are significantly higher than others. Specifically, our model achieves a performance gain of 1.88 dB over the recent WeatherDiff \cite{ozdenizci2023restoring} in PSNR indicator.

\noindent\textbf{Rain-by-raindrop Removal} For restoring vision in rain-by-raindrop weather, we evaluate the effectiveness of comparative methods on RainDS-syn dataset. As Table \ref{tab:specific}(h) depicted, the quantitative performance of our model is better than IDT \cite{xiao2022image}.

\begin{figure*}[!t]
\begin{minipage}{\linewidth}
    \centering
    \includegraphics[width=2.9cm]{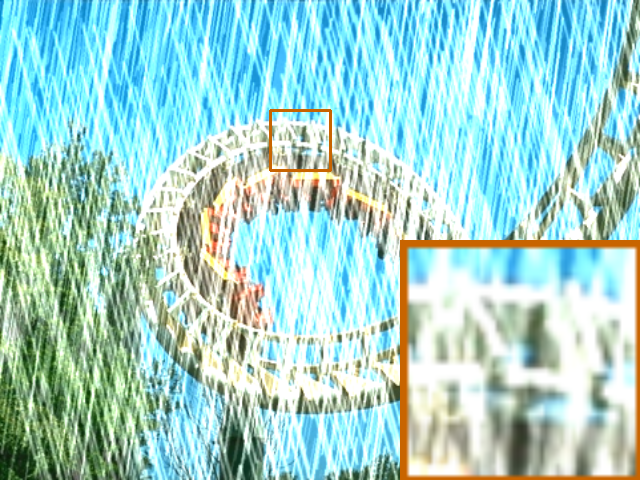}
    \includegraphics[width=2.9cm]{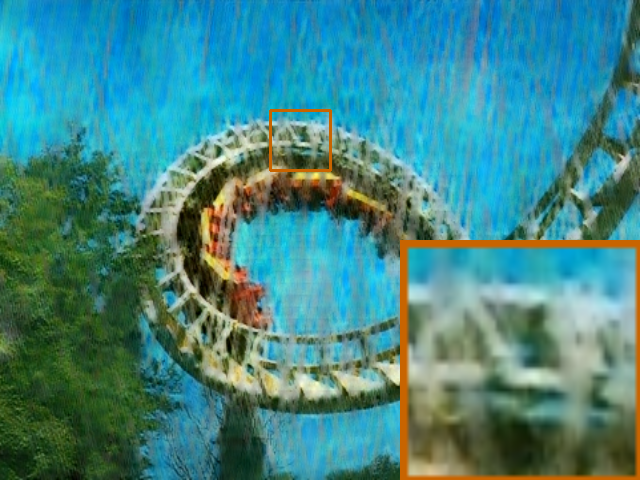}
    \includegraphics[width=2.9cm]{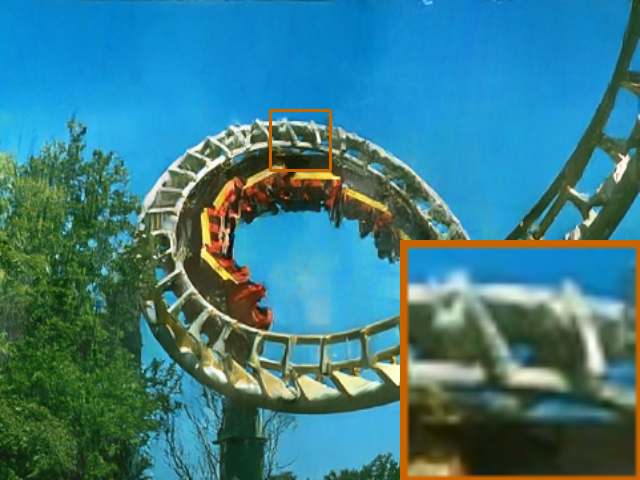}
    \includegraphics[width=2.9cm]{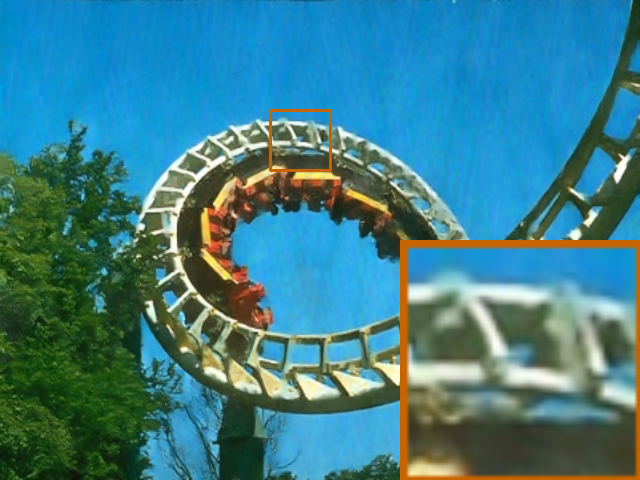}
    \includegraphics[width=2.9cm]{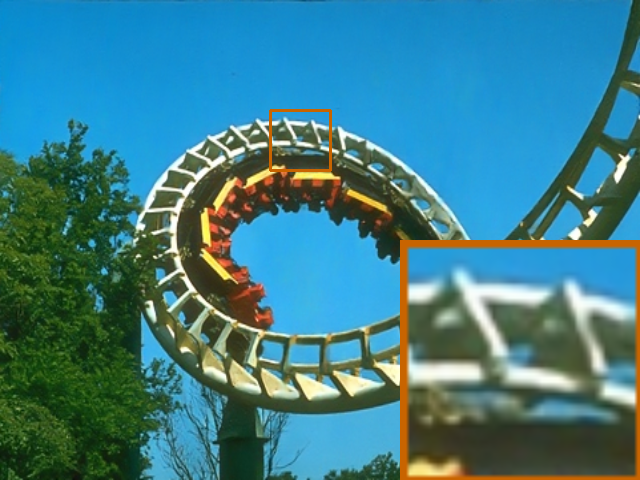}
    \includegraphics[width=2.9cm]{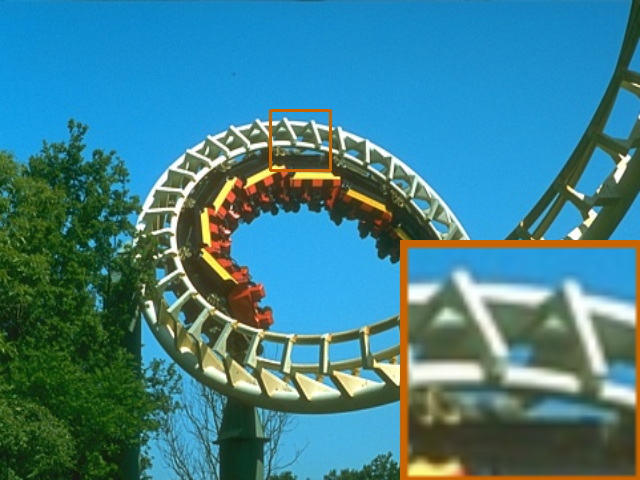}

    \footnotesize
    \makebox[2.9cm][c]{Input Image}
    \makebox[2.9cm][c]{AIRFormer \cite{gao2023frequency}}
    \makebox[2.9cm][c]{WeatherDiff \cite{ozdenizci2023restoring}}
    \makebox[2.9cm][c]{PromptIR \cite{potlapalli2023promptir}}
    \makebox[2.9cm][c]{\textbf{ADSM (ours)}}
    \makebox[2.9cm][c]{Ground Truth}
    
    \caption{Visual comparisons of the evaluated methods on all-in-one image deraining, where colored boxes correspond to the zoomed-in patch for better comparisons. Our proposed approach achieves the more visual pleasing deraining results over the other evaluated methods.}
    \label{fig:rain}
    \end{minipage}
\hfill
\begin{minipage}{\linewidth}
    \centering
    \vspace{2mm}
    
    \includegraphics[width=2.9cm]{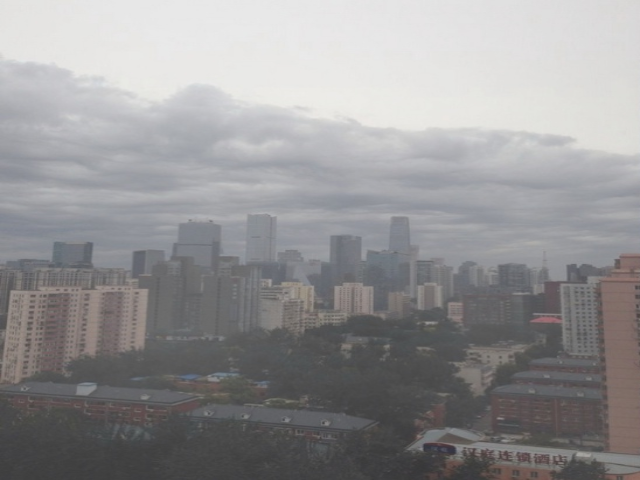}
    \includegraphics[width=2.9cm]{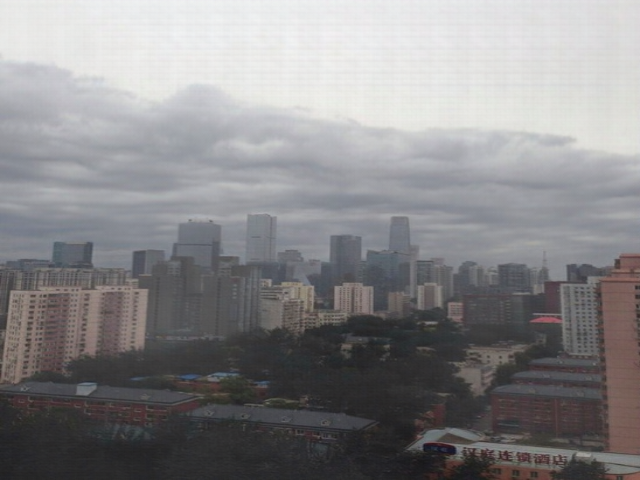}
    \includegraphics[width=2.9cm]{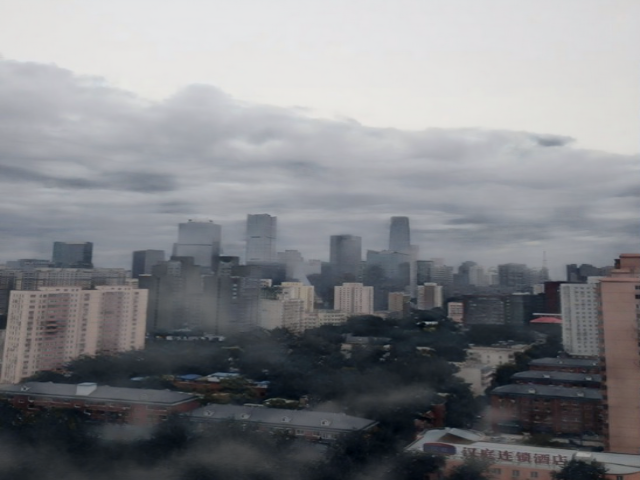}
    \includegraphics[width=2.9cm]{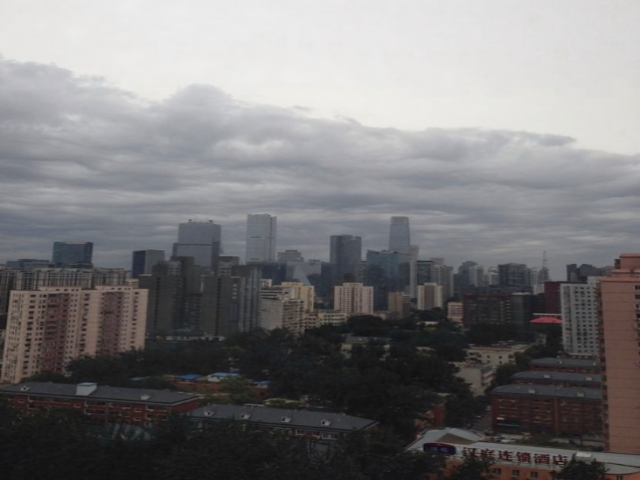}
    \includegraphics[width=2.9cm]{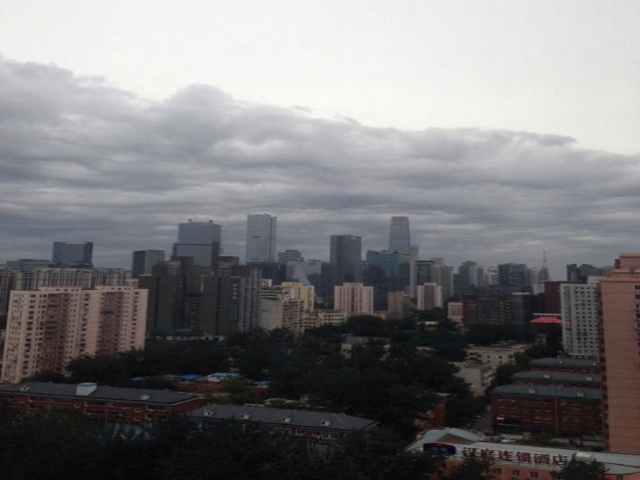}
    \includegraphics[width=2.9cm]{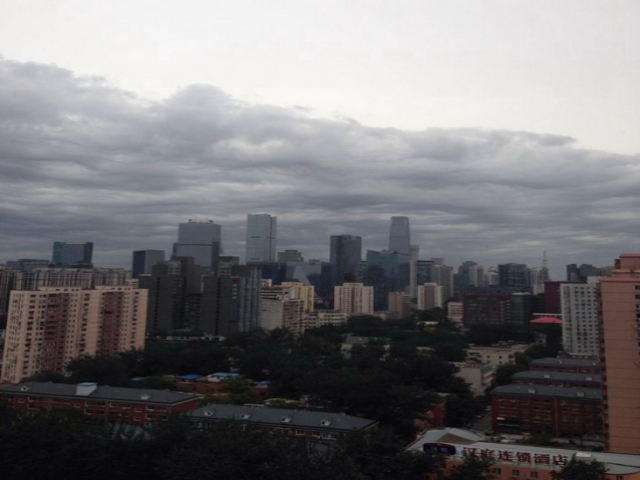}

    \vspace{1mm}
    \includegraphics[width=2.9cm]{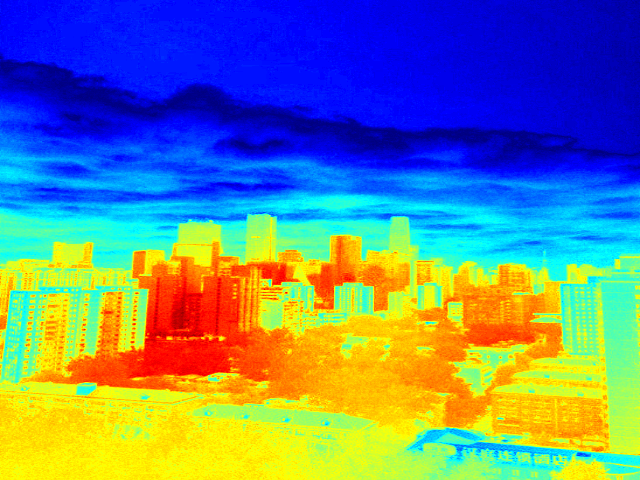}
    \includegraphics[width=2.9cm]{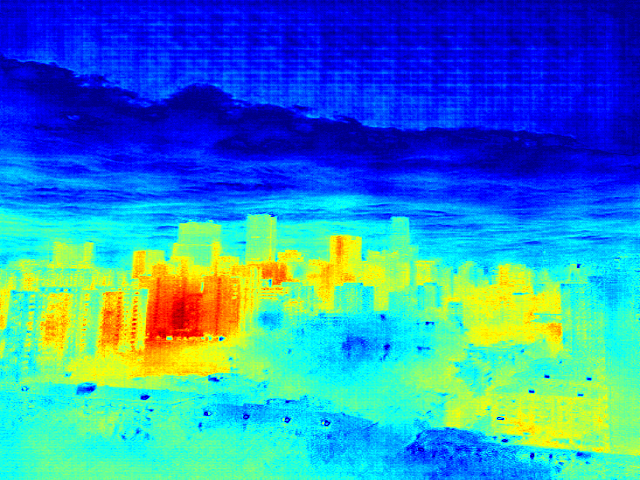}
    \includegraphics[width=2.9cm]{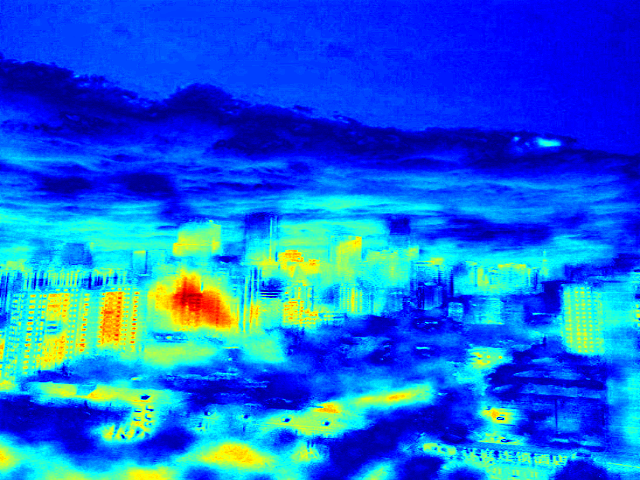}
    \includegraphics[width=2.9cm]{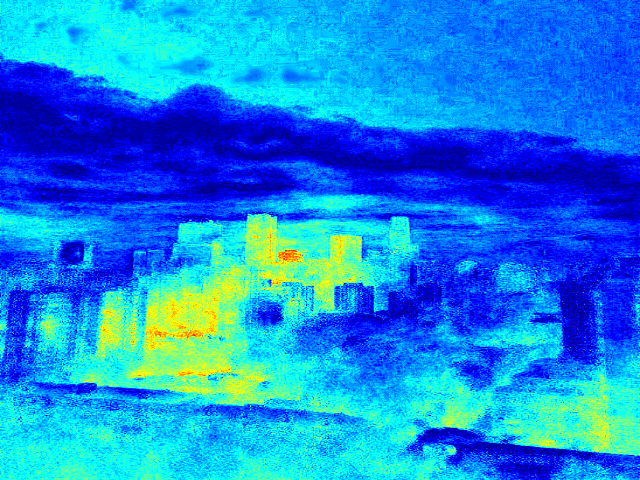}
    \includegraphics[width=2.9cm]{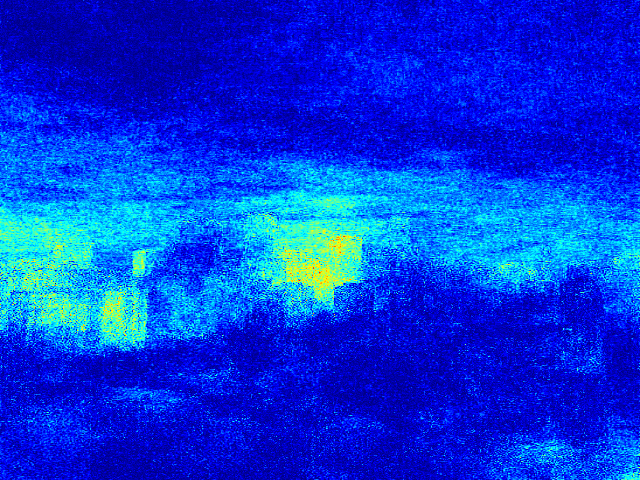}
    \includegraphics[width=2.9cm]{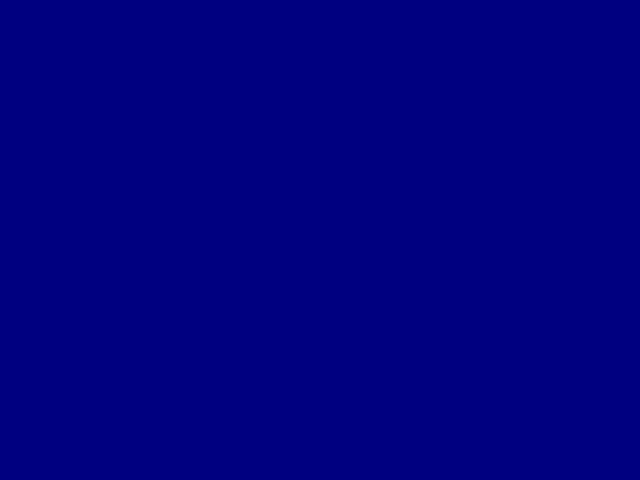}

    \footnotesize
    \makebox[2.9cm][c]{Input Image}
    \makebox[2.9cm][c]{AIRFormer \cite{gao2023frequency}}
    \makebox[2.9cm][c]{WeatherDiff \cite{ozdenizci2023restoring}}
    \makebox[2.9cm][c]{PromptIR \cite{potlapalli2023promptir}}
    \makebox[2.9cm][c]{\textbf{ADSM (ours)}}
    \makebox[2.9cm][c]{Ground Truth}
    
    \caption{Visual comparisons of the evaluated methods on all-in-one image dehazing, where below the reconstructed images are the error maps between them and the corresponding clean images. The darker the error maps, the better the quality of the reconstructed images. Our proposed approach achieves the more visual pleasing dehazing results over the other evaluated methods.}
    \label{fig:haze}
    \end{minipage}
\hfill
\begin{minipage}{\linewidth}
    \centering
    \vspace{2mm}
    \includegraphics[width=2.9cm]{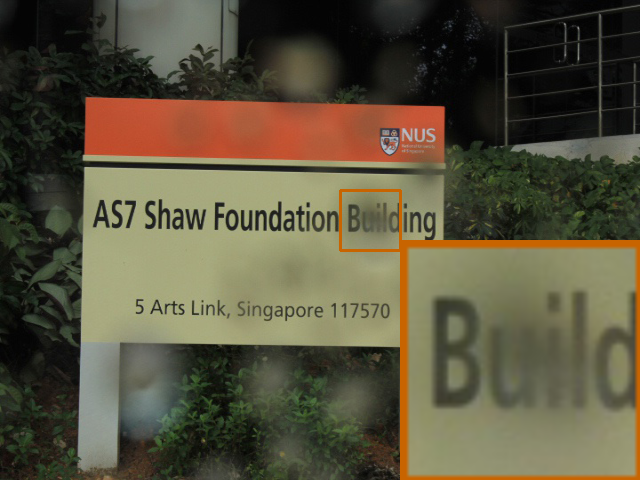}
    \includegraphics[width=2.9cm]{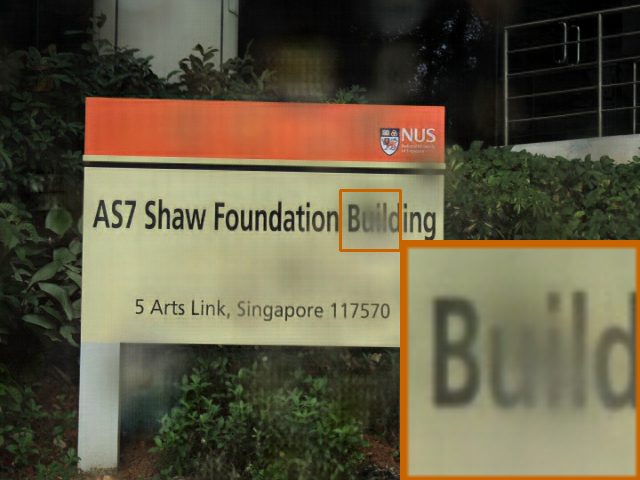}
    \includegraphics[width=2.9cm]{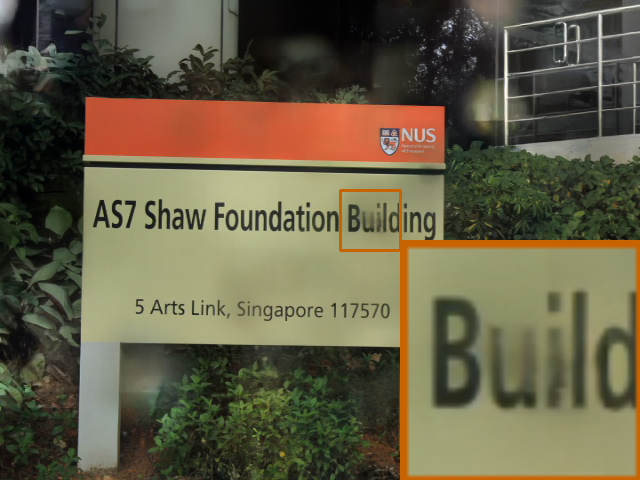}
    \includegraphics[width=2.9cm]{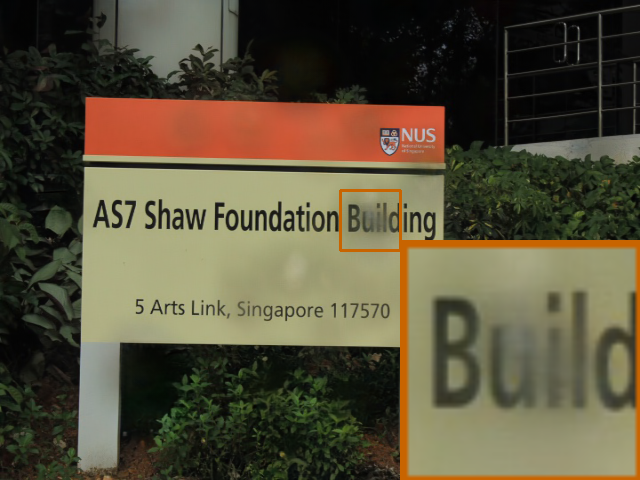}
    \includegraphics[width=2.9cm]{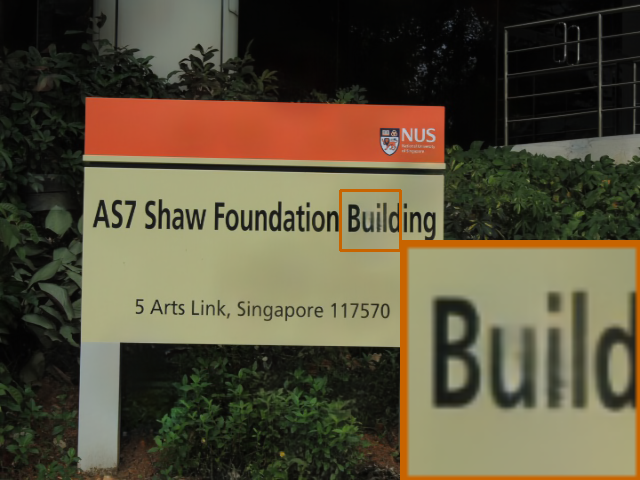}
    \includegraphics[width=2.9cm]{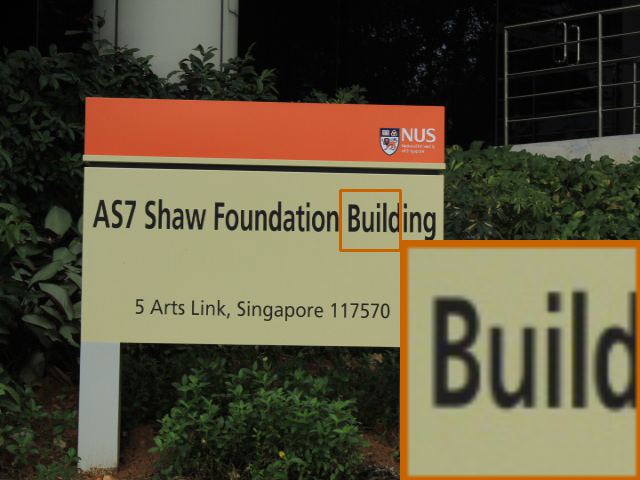}

    \footnotesize
    \makebox[2.9cm][c]{Input Image}
    \makebox[2.9cm][c]{AIRFormer \cite{gao2023frequency}}
    \makebox[2.9cm][c]{WeatherDiff \cite{ozdenizci2023restoring}}
    \makebox[2.9cm][c]{PromptIR \cite{potlapalli2023promptir}}
    \makebox[2.9cm][c]{\textbf{ADSM (ours)}}
    \makebox[2.9cm][c]{Ground Truth}
    
    \caption{Visual comparisons of the evaluated methods on all-in-one raindrop removal, where colored boxes correspond to the zoomed-in patch for better comparisons. Our proposed approach achieves the more visual pleasing raindrop removal results over the other evaluated methods.}
    \label{fig:raindrop}
    \end{minipage}
    \hfill
    \begin{minipage}{\linewidth}
    \centering
    \includegraphics[width=2.9cm]{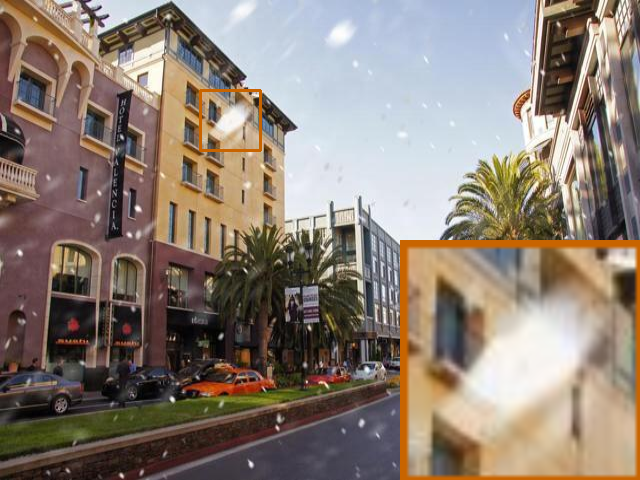}
    \includegraphics[width=2.9cm]{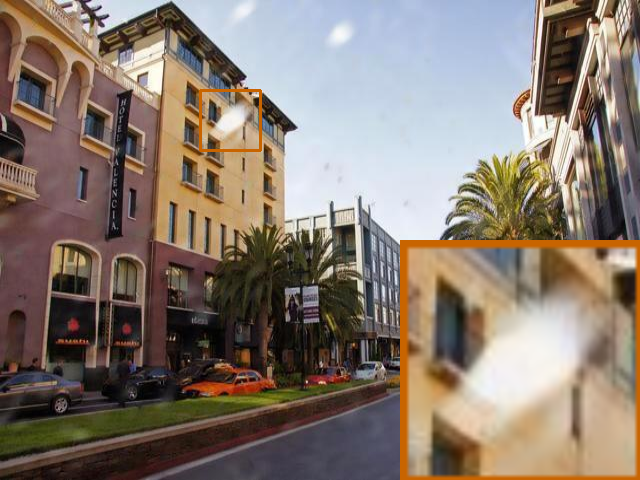}
    \includegraphics[width=2.9cm]{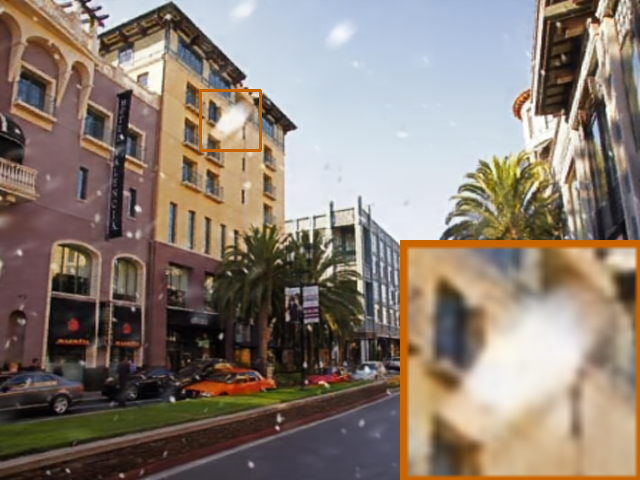}
    \includegraphics[width=2.9cm]{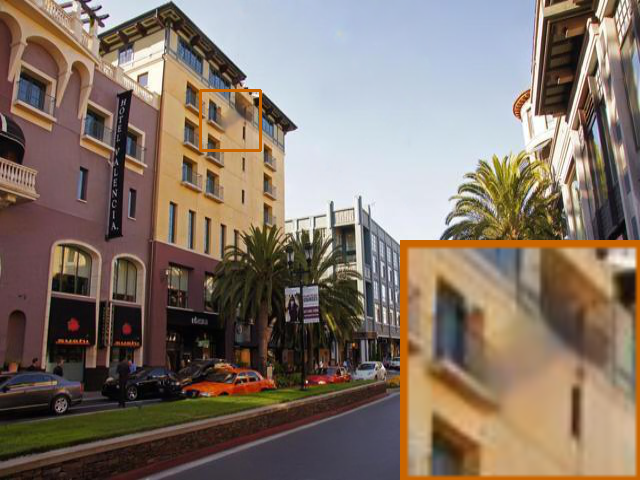}
    \includegraphics[width=2.9cm]{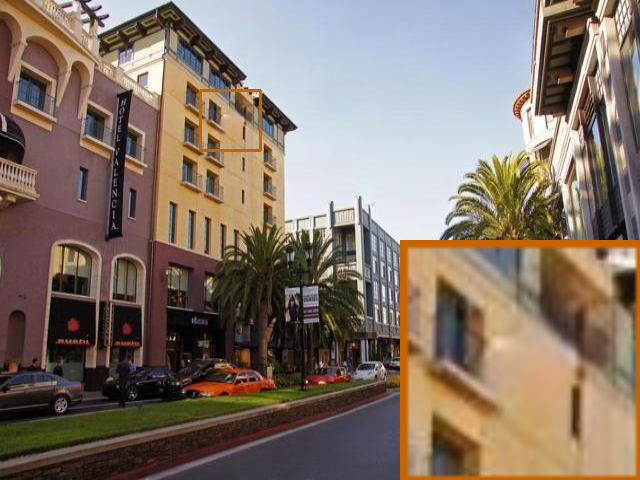}
    \includegraphics[width=2.9cm]{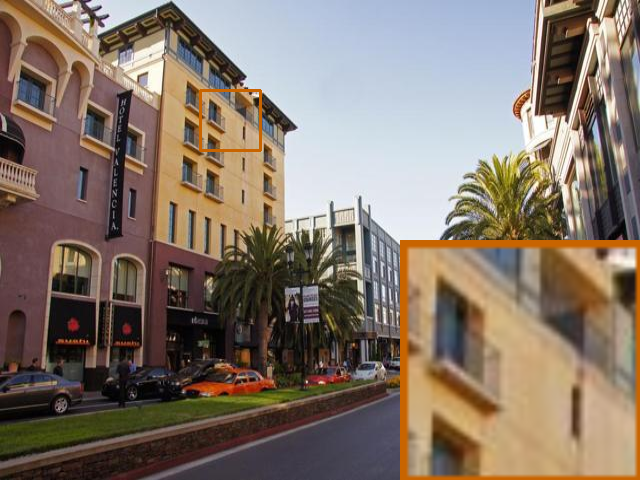}

    \footnotesize
    \makebox[2.9cm][c]{Input Image}
    \makebox[2.9cm][c]{AIRFormer \cite{gao2023frequency}}
    \makebox[2.9cm][c]{WeatherDiff \cite{ozdenizci2023restoring}}
    \makebox[2.9cm][c]{PromptIR \cite{potlapalli2023promptir}}
    \makebox[2.9cm][c]{\textbf{ADSM (ours)}}
    \makebox[2.9cm][c]{Ground Truth}
    
    \caption{Visual comparisons of the evaluated methods on all-in-one image desnowing, where colored boxes correspond to the zoomed-in patch for better comparisons. Our proposed approach achieves the more visual pleasing desnowing results over the other evaluated methods.}
    \label{fig:snow}
    \end{minipage}
    \hfill
    \begin{minipage}{\linewidth}
    \centering
    \vspace{2mm}
    \includegraphics[width=2.9cm]{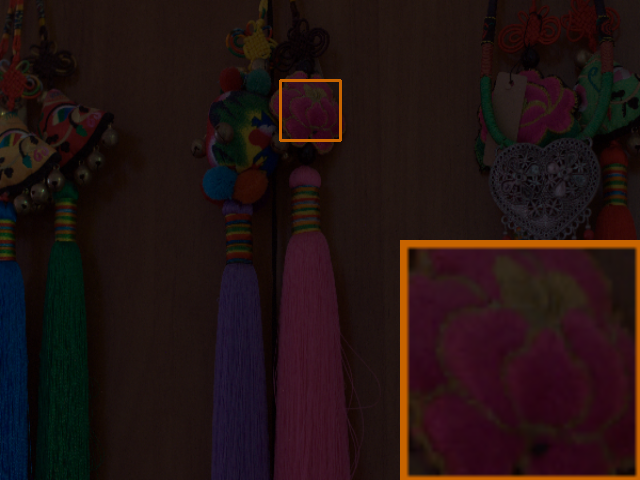}
    \includegraphics[width=2.9cm]{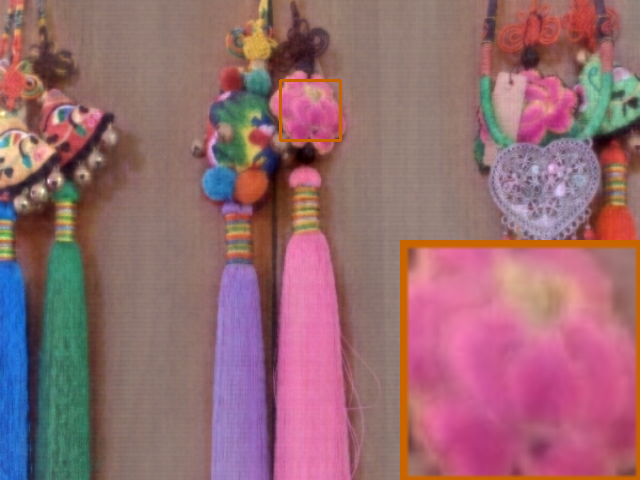}
    \includegraphics[width=2.9cm]{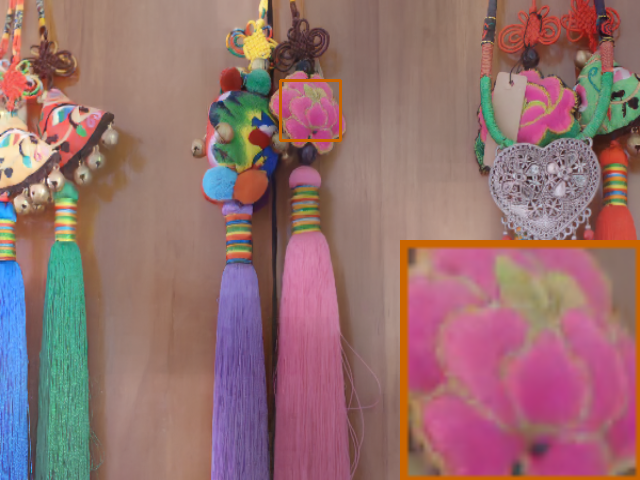}
    \includegraphics[width=2.9cm]{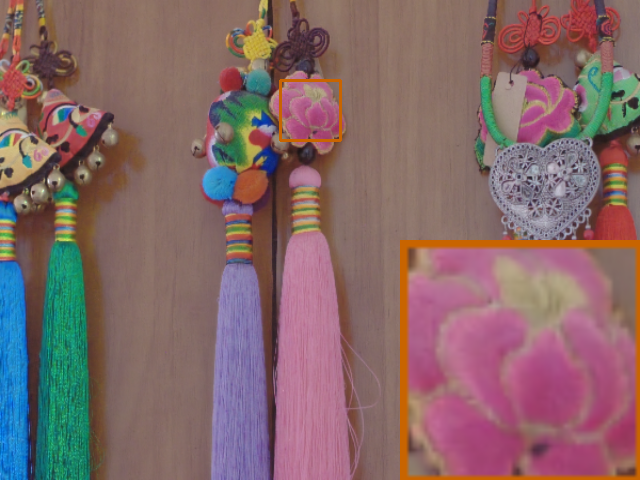}
    \includegraphics[width=2.9cm]{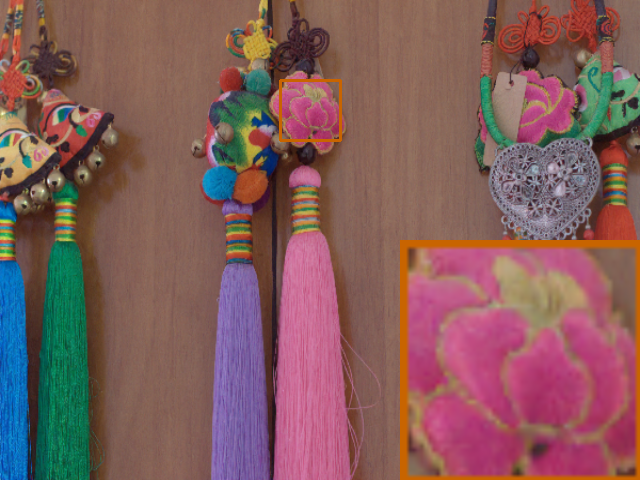}
    \includegraphics[width=2.9cm]{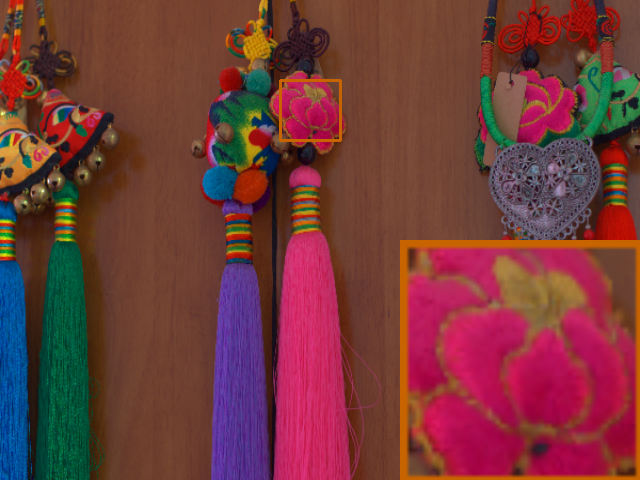}

    \footnotesize
    \makebox[2.9cm][c]{Input Image}
    \makebox[2.9cm][c]{AIRFormer \cite{gao2023frequency}}
    \makebox[2.9cm][c]{WeatherDiff \cite{ozdenizci2023restoring}}
    \makebox[2.9cm][c]{PromptIR \cite{potlapalli2023promptir}}
    \makebox[2.9cm][c]{\textbf{ADSM (ours)}}
    \makebox[2.9cm][c]{Ground Truth}
    
    \caption{Visual comparisons of the evaluated methods on all-in-one image enhancement, where colored boxes correspond to the zoomed-in patch for better comparisons. Our proposed approach achieves the more visual pleasing enhancement results over the other evaluated methods.}
    \label{fig:low-light}
    \end{minipage}
\hfill
\end{figure*}

\begin{figure*}[htp]
\begin{minipage}{\linewidth}
    \centering
    \vspace{2mm}
    \includegraphics[width=2.9cm]{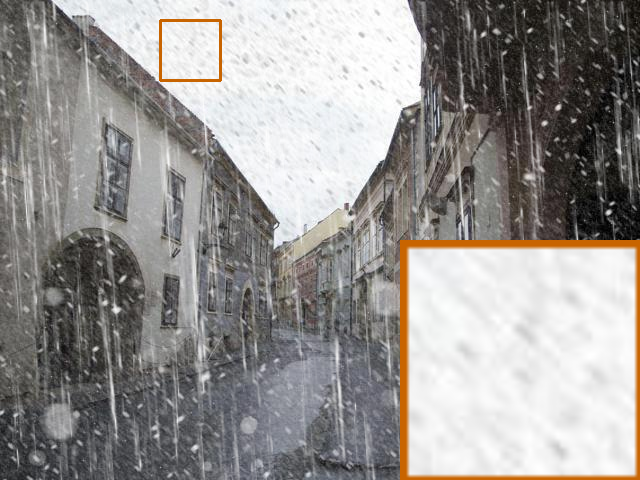}
    \includegraphics[width=2.9cm]{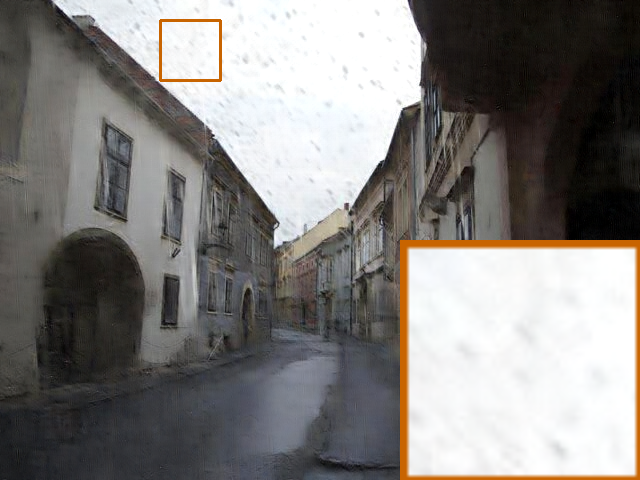}
    \includegraphics[width=2.9cm]{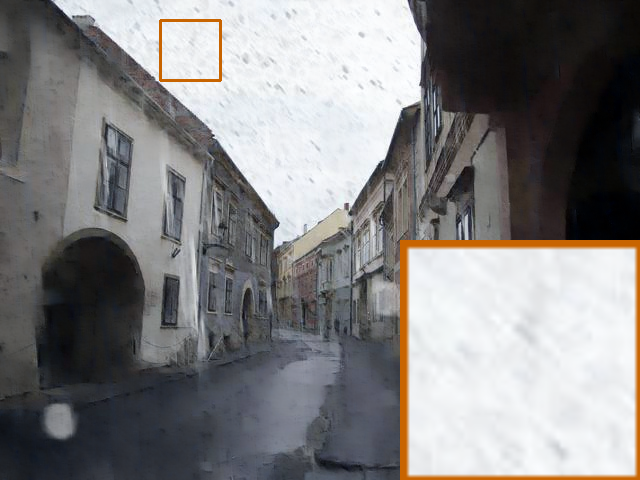}
    \includegraphics[width=2.9cm]{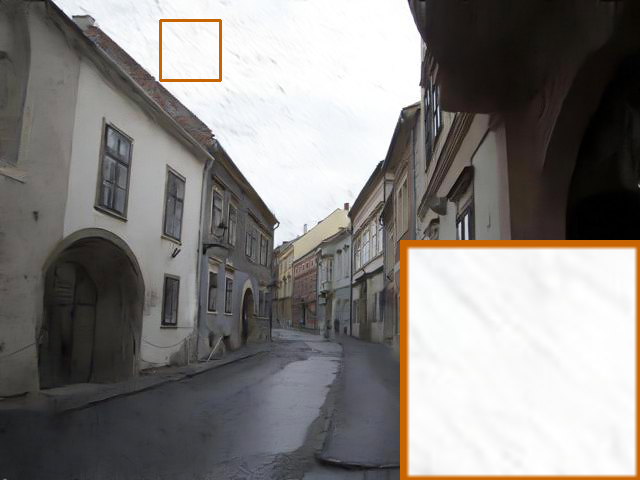}
    \includegraphics[width=2.9cm]{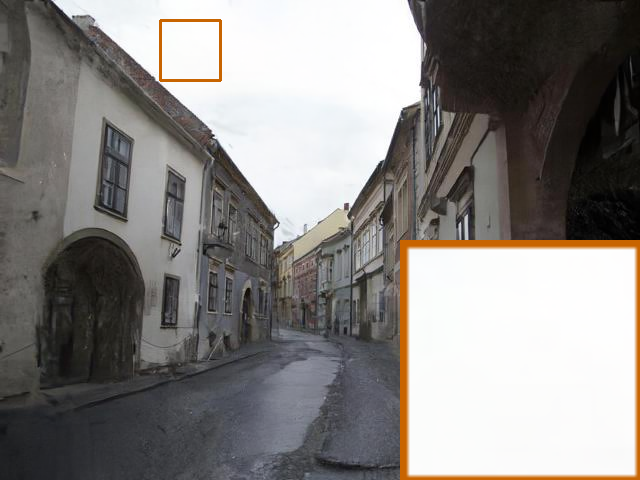}
    \includegraphics[width=2.9cm]{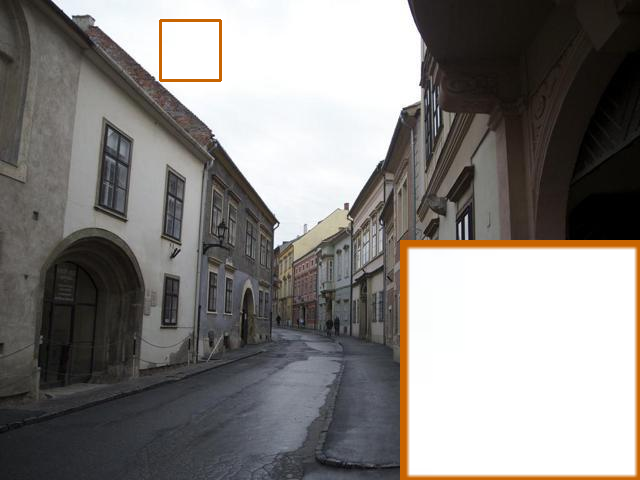}

    \footnotesize
    \makebox[2.9cm][c]{Input Image}
    \makebox[2.9cm][c]{AIRFormer \cite{gao2023frequency}}
    \makebox[2.9cm][c]{WeatherDiff \cite{ozdenizci2023restoring}}
    \makebox[2.9cm][c]{PromptIR \cite{potlapalli2023promptir}}
    \makebox[2.9cm][c]{\textbf{ADSM (ours)}}
    \makebox[2.9cm][c]{Ground Truth}
    
    \caption{Visual comparisons of the evaluated methods on all-in-one rain-by-snow removal, where colored boxes correspond to the zoomed-in patch for better comparisons. Our proposed approach achieves the more visual pleasing rain-by-snow removal results over the other evaluated methods.}
    \label{fig:rain-by-snow}
    \end{minipage}
    \hfill
    \begin{minipage}{\linewidth}
    \centering
    \includegraphics[width=2.9cm]{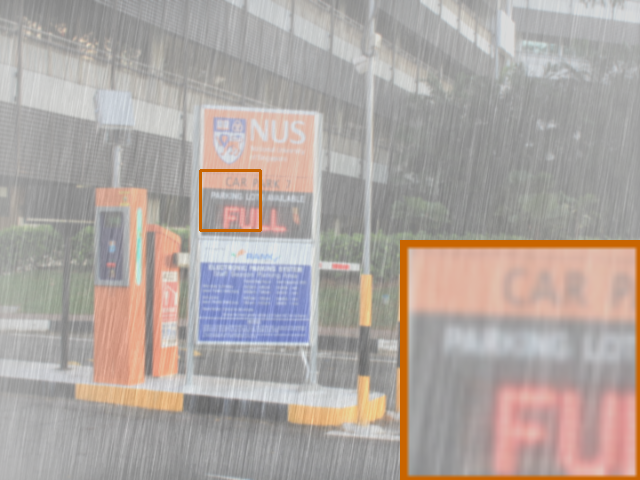}
    \includegraphics[width=2.9cm]{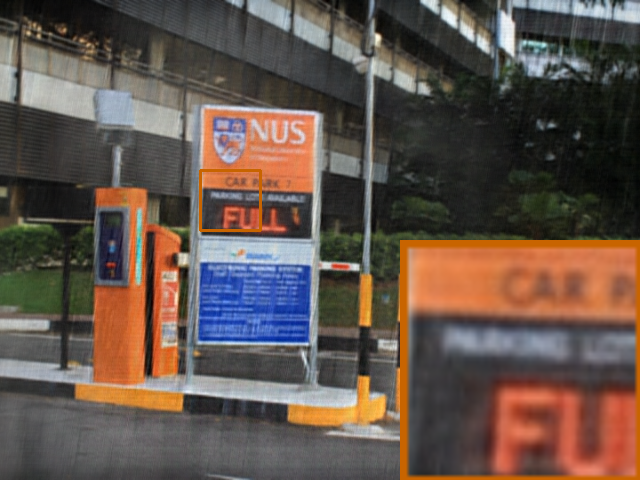}
    \includegraphics[width=2.9cm]{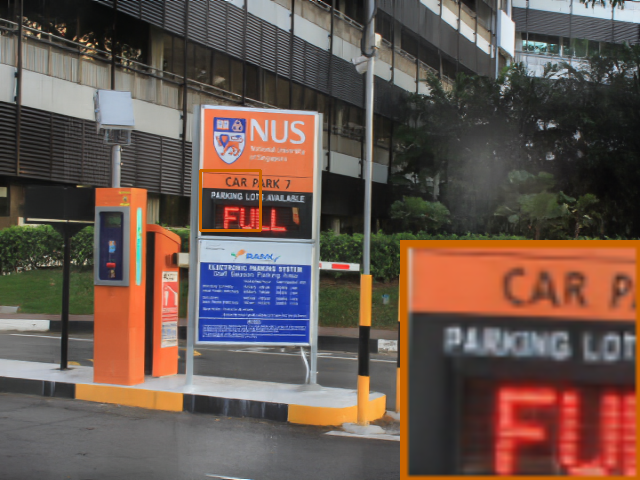}
    \includegraphics[width=2.9cm]{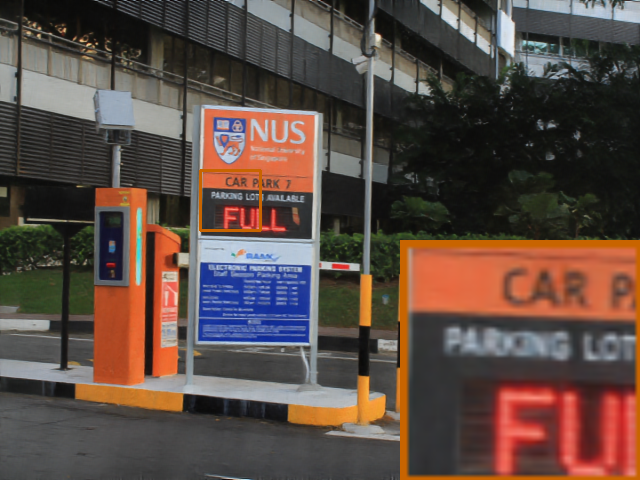}
    \includegraphics[width=2.9cm]{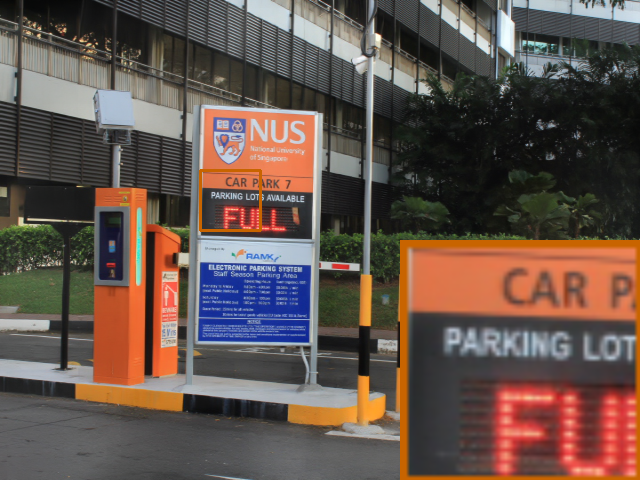}
    \includegraphics[width=2.9cm]{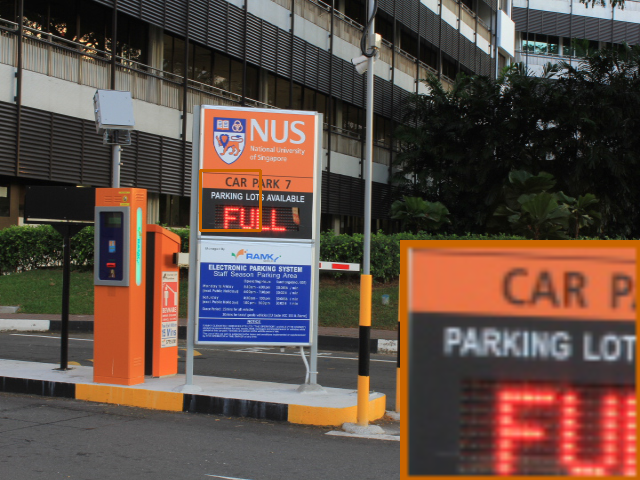}

    \footnotesize
    \makebox[2.9cm][c]{Input Image}
    \makebox[2.9cm][c]{AIRFormer \cite{gao2023frequency}}
    \makebox[2.9cm][c]{WeatherDiff \cite{ozdenizci2023restoring}}
    \makebox[2.9cm][c]{PromptIR \cite{potlapalli2023promptir}}
    \makebox[2.9cm][c]{\textbf{ADSM (ours)}}
    \makebox[2.9cm][c]{Ground Truth}
    
    \caption{Visual comparisons of the evaluated methods on all-in-one rain-by-haze removal, where colored boxes correspond to the zoomed-in patch for better comparisons. Our proposed approach achieves the more visual pleasing rain-by-haze removal results over the other evaluated methods.}
    \label{fig:rain-by-haze}
    \end{minipage}
    \hfill
    \begin{minipage}{\linewidth}
    \centering
    \vspace{2mm}
    \includegraphics[width=2.9cm]{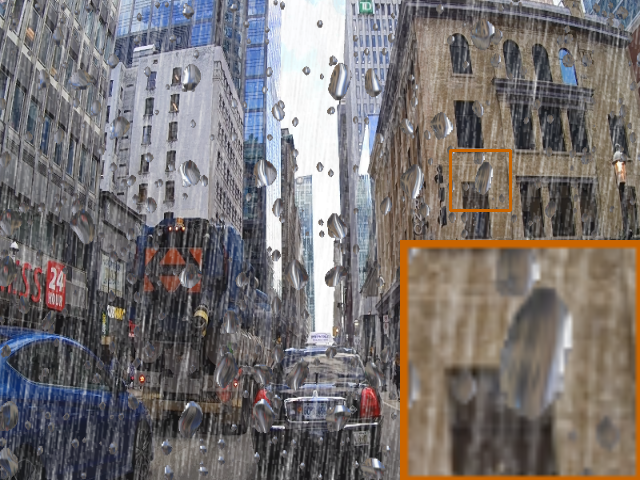}
    \includegraphics[width=2.9cm]{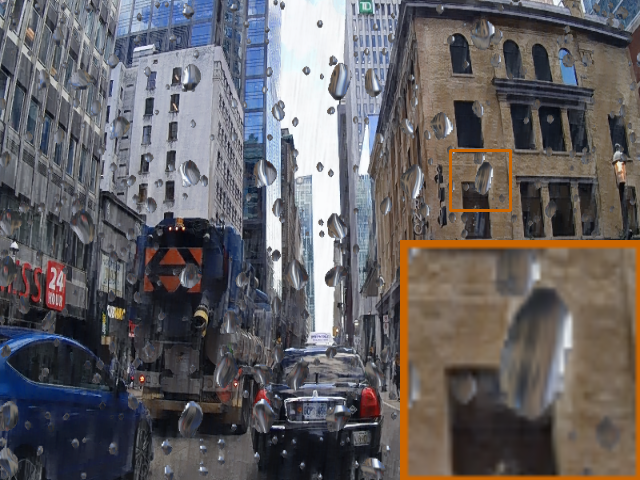}
    \includegraphics[width=2.9cm]{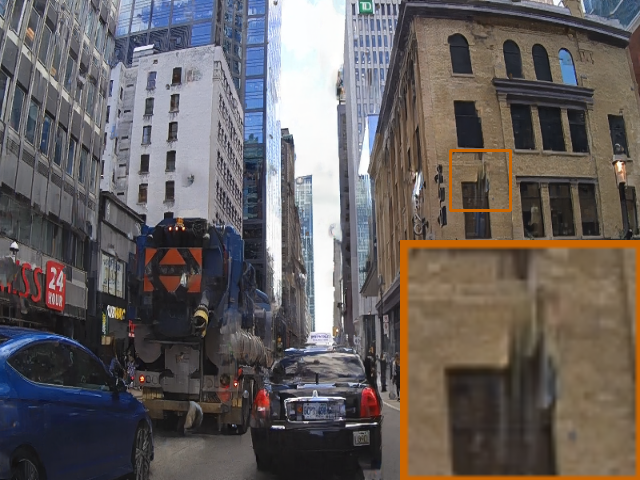}
    \includegraphics[width=2.9cm]{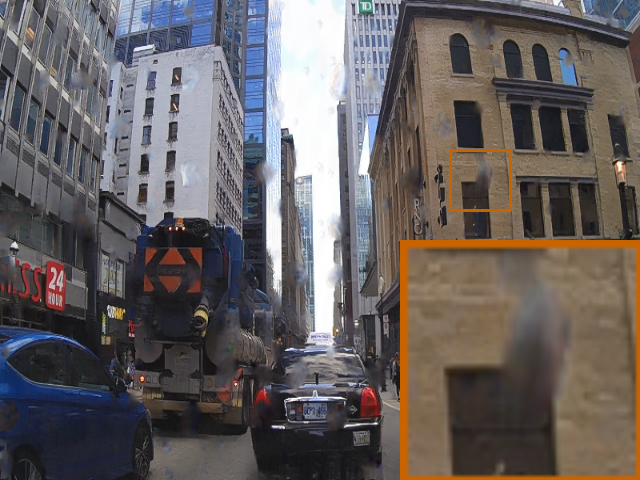}
    \includegraphics[width=2.9cm]{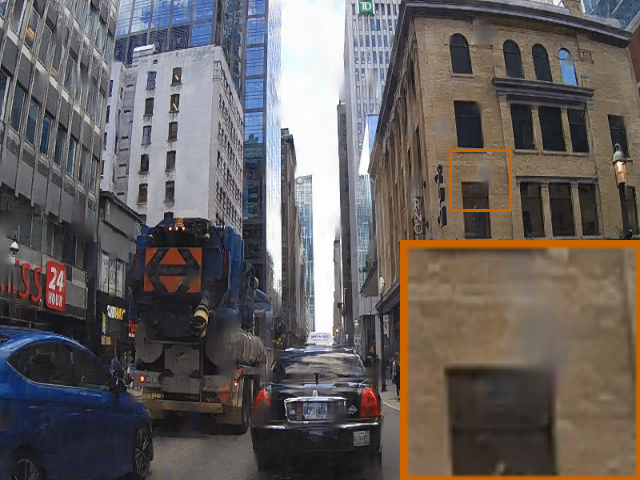}
    \includegraphics[width=2.9cm]{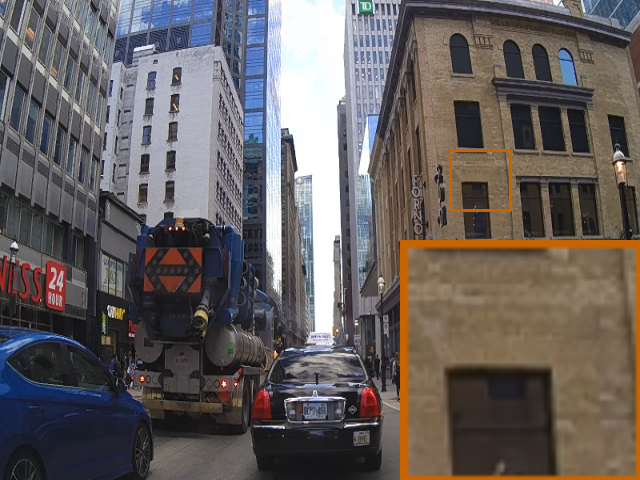}

    \footnotesize
    \makebox[2.9cm][c]{Input Image}
    \makebox[2.9cm][c]{AIRFormer \cite{gao2023frequency}}
    \makebox[2.9cm][c]{WeatherDiff \cite{ozdenizci2023restoring}}
    \makebox[2.9cm][c]{PromptIR \cite{potlapalli2023promptir}}
    \makebox[2.9cm][c]{\textbf{ADSM (ours)}}
    \makebox[2.9cm][c]{Ground Truth}
    
    \caption{Visual comparisons of the evaluated methods on all-in-one rain-by-raindrop removal, where colored boxes correspond to the zoomed-in patch for better comparisons. Our proposed approach achieves the more visual pleasing rain-by-raindrop removal results over the other evaluated methods.}
    \label{fig:rain-by-raindrop}
    \end{minipage}
\end{figure*}

\subsection{Comparisons with All-in-one Setting}

We also evaluate our model as all-in-one method for all-in-one weather-degraded image restoration. In this setting, we train seven task-specific methods, six all-in-one methods and our proposed approach on AWIR120K dataset and conduct validations on the corresponding testing dataset.

\noindent\textbf{Image Deraining}
We report the quantitative deraining performance of our proposed ADSM in Table \ref{tab:all-in-one}(a). As illustrated, our model advances the second-best PromptIR \cite{potlapalli2023promptir} by 0.45 dB in PSNR. Meanwhile, our model also achieves the best performance in SSIM indicator. For better comparisons, we include the visual comparisons of the evaluated methods in Figure \ref{fig:rain}. As depicted, the reconstructed images obtained by our approach show better textural details.

\noindent\textbf{Image Dehazing}
Table \ref{tab:all-in-one}(b) illustrates the quantitative dehazing comparisons of the evaluated methods. As reported in this table, our proposed method obtains a performance gain of 0.98 dB over promptIR \cite{potlapalli2023promptir} in PSNR. Meanwhile, we present the visual samples and error maps of the dehazed results generated by our model and other methods in Figure \ref{fig:haze}. As shown, our proposed approach generates the reconstructed images that are closer to the corresponding clean images.

\noindent\textbf{Raindrop Removal}
We conduct comparative experiments to evaluated the raindrop removal performance of our proposed ADSM. As Table \ref{tab:all-in-one}(c) reported, our method achieves substantial improvements in both PSNR and SSIM indicators over the other evaluated methods.
Specifically, our model achieves a performance gain of 0.63 dB over PromptIR \cite{potlapalli2023promptir}.
To further demonstrate the advantages of our model, we depict the visual comparisons of the evaluated methods in Figure \ref{fig:raindrop}.
As illustrated, AIRFormer \cite{gao2023frequency} does not effectively remove raindrop degradation, and the other two comparative methods result in significant color distortions in the restored images. In contrast, our proposed method successfully eliminates raindrops and achieves reconstructed images that closely resemble the corresponding ground truths.

\noindent\textbf{Image Desnowing}
To evaluate the desnowing performance of our proposed ADSM, we conduct the validations of the evaluated methods on Snow100K-L dataset \cite{liu2018desnownet}. We include the quantitative comparisons in Table \ref{tab:all-in-one}(d), where our model achieves the best performance in both PSNR and SSIM indicators.
Meanwhile, we present the visual comparisons of different methods in Figure \ref{fig:snow}.
As depicted, AIRFormer \cite{gao2023frequency} and WeatherDiff \cite{ozdenizci2023restoring} fail to eliminate the snow particles, while PromptIR \cite{potlapalli2023promptir} removes most of the snow particles, but there are degradation residues in the generated images.
In contrast, our model achieves the best performance in eliminating the snow degradation.

\noindent\textbf{Image Enhancement}
For image enhancement, we conduct comparative experiments of the evaluated methods and our proposed ADSM. As Table \ref{tab:all-in-one}(e) reported, our proposed method achieves a performance improvement by 1.06 dB over DehazeFormer \cite{song2023vision}.
Meanwhile, we illustrate the visual comparisons of the involved methods in Figure \ref{fig:low-light}.
As depicted, the enhanced images generated by the other methods appear severe color distortions.
However, our model successfully preserves the color fidelity and the details of reconstructed images.

\noindent\textbf{Rain-by-snow Removal}
Table \ref{tab:all-in-one}(f) illustrates the quantitative comparisons of our proposed ADSM and the other comparative methods on rain-by-snow weather removal.
As reported in Table \ref{tab:all-in-one}(f), PromptIR \cite{potlapalli2023promptir} obtains the second-best performance in PSNR, while our model achieves a performance gain of 0.53 dB.
The visual samples processed by the involved methods are shown in Figure \ref{fig:rain-by-snow}.
As depicted, all comparative methods can effectively eliminate rain streaks.
However, when snow particles are also present simultaneously, these other methods fail to remove these imperfections. In contrast, our model excels in removing snow particles, producing results with the highest color fidelity.

\noindent\textbf{Rain-by-haze Removal}
We illustrate the quantitative performance of our proposed method on rain-by-haze weather removal in Table \ref{tab:all-in-one}(g), where our approach achieves the best metrical scores.
Specifically, our model obtains a performance gain of 0.72 dB in PSNR over PromptIR \cite{potlapalli2023promptir}.
Meanwhile, we also provide the visual comparisons in Figure \ref{fig:rain-by-haze}.
As illustrated, all comparative methods effectively remove the coexisting rain streaks and haze veiling, however the details obtained by our model are closer to the corresponding ground truths.

\noindent\textbf{Rain-by-raindrop Removal}
Table \ref{tab:all-in-one}(h) reports the quantitative performance of the comparative methods for rain-by-raindrop weather removal.
As illustrated, our model advances PromptIR \cite{potlapalli2023promptir} by 0.37 dB in PSNR indicator.
Figure \ref{fig:rain-by-raindrop} shows that AIRFormer \cite{gao2023frequency} is unable to remove raindrops.
Meanwhile, the reconstructed results obtained by WeatherDiff \cite{ozdenizci2023restoring} and PromptIR \cite{potlapalli2023promptir} suffer from severe degradation residues.
However, our proposed model completely eliminates the co-existing rain streaks and raindrops, while preserving the image contents.

\subsection{Ablation Studies}

We conduct ablation experiments to evaluate how well our suggested latent prompt generators and wavelet-based noise estimation network perform. We randomly pick ten pairs of images from each of the eight weather-degraded datasets to create a small testing dataset for our ablation validation.

\noindent\textbf{Latent Prompt Generators}
We present the results of ablative experiments on the latent prompt generators in Table \ref{tab:lp}.
In this table, w/ and w/o indicate $'with'$ and $'without'$, respectively. As shown, utilizing the pre-trained latent prompt generators leads to a significant performance improvement of 3.93 dB compared to the model without our proposed LPGs. Therefore, LPGs guide the reverse sampling process to mitigate multiple degradations and accurately reconstruct image contents, thereby enhancing the performance in handling various types of degradations.
In contrast, the absence of pre-training in the latent prompt generators results in a noticeable performance decline for our proposed ADSM, which underscores the importance of pre-training LPGs in generating accurate latent prompts.
Furthermore, removing the caption prompt generator causes an overall performance decrease of 2.76 dB, highlighting its role in accurately reconstructing image contents. Overall, all three latent prompt generators prove beneficial in enhancing the performance of our model.

\begin{table}[h]\footnotesize
    \centering
    \caption{Ablation experiments on our proposed latent prompt generators. The latent prompts of degradation type, degradation property and image caption present positive effects in the overall performance for all-in-one weather-degraded image restoration.}
    \label{tab:lp}
    \renewcommand\arraystretch{1.15}
    \setlength{\tabcolsep}{5.5mm}{
    \begin{tabular}{llcc}
    \toprule
    \multicolumn{2}{c}{\hspace{-10mm}Model} & PSNR & SSIM \\
    \midrule
    w/o & latent promptors & 26.31 & 0.877 \\
    w/o & pre-trained & 16.85 & 0.698 \\
    w/ & pre-trained & \red{\textbf{30.24}} & \red{\textbf{0.929}} \\
    w/o & caption promptor & \blue{\textbf{27.48}} & \blue{\textbf{0.883}} \\
    \bottomrule
    \end{tabular}
    }
\end{table}

\noindent\textbf{Noise Estimating Network}
We also conduct individual assessments to gauge how well our proposed modules perform within our WNE-Net. As shown in Table \ref{tab:WNE-Net}, replacing our WSRB with the commonly used residual block results in an overall performance decrease of 1.13 dB in PSNR.
Additionally, substituting our WFSB with convolution for cross-stage sampling leads to a performance drop of 1.38 dB. Therefore, incorporating discrete wavelet transform into the noise estimating network proves effective for accurate noise estimation.

\begin{table}[h]\footnotesize
    \centering
    \caption{Ablation experiments on noise estimating network. Each proposed component plays a valuable role on the overall performance for noise estimation.}
    \label{tab:WNE-Net}
    \renewcommand\arraystretch{1.15}
    \setlength{\tabcolsep}{7mm}{
    \begin{tabular}{llcc}
    \toprule
    \multicolumn{2}{c}{\hspace{-2mm}Model} & PSNR & SSIM \\
    \midrule
    w/o & WSRB & \blue{\textbf{29.11}} & \blue{\textbf{0.915}} \\
    w/o & WFSB & 28.86 & 0.902 \\
    \multicolumn{2}{c}{\textbf{WNE-Net}} & \red{\textbf{30.24}} & \red{\textbf{0.929}} \\
    \bottomrule
    \end{tabular}
    }
\end{table}

\noindent\textbf{Complexities of self-attention mechanisms}
To verify the capability of wavelet transform in reducing the instantaneous computational burden during self-attention representation, we compare the computational cost of traditional self-attention mechanism and wavelet self-attention mechanism. The input feature structure we selected is (32, 64, 64), and the feature structure became (128, 32, 32) after wavelet transform. The comparisons of computational cost is shown in Table \ref{tab:computation}, where the computational burden for representing self-attention with wavelet-transformed features decreases by an order of magnitude in both addition and multiplication calculations. Therefore, using wavelet transform for self-attention representation not only reduces the computational burden but also improves the ability of noise estimating network to perceive noise in the frequency domain.

\begin{table}[h]\footnotesize
    \centering
    \caption{Comparisons on the computational amount of different self-attention mechanisms. The computational burden for representing self-attention with wavelet-transformed features is reduced by an order of magnitude in terms of both addition and multiplication operations.}
    \label{tab:computation}
    \renewcommand\arraystretch{1.15}
    \setlength{\tabcolsep}{7mm}{
    \begin{tabular}{lcc}
    \toprule
    Attention & Addition & Multiplication \\
    \midrule
    Traditional & \blue{$\mathbf{1.07\times 10^9}$} & \blue{$\mathbf{1.07\times 10^9}$} \\
    Wavelet & \red{$\mathbf{2.69\times 10^8}$} & \red{$\mathbf{2.68\times 10^8}$} \\
    \bottomrule
    \end{tabular}
    }
\end{table}

\section{Limitations}
Although our solution demonstrates superiority over existing all-in-one models, it still has some limitations. Primarily, the fidelity of image reconstruction heavily depends on the prompts generated by our latent prompt generators. This necessitates that our generators accurately characterize the type and property of degradation, as well as the image captions. Additionally, the significant inference time and computational burden of the diffusion model severely restrict the real-world applicability of our approach.

\section{Conclusion}

In this work, we introduce an adaptive degradation-aware self-prompting model for all-in-one weather-degraded image restoration.
Initially, latent prompt generators are utilized to derive prompts corresponding to specific degradation types, degradation properties and image captions. These prompts enable our restoration model to understand prevalent weather distortions and effectively aid in reconstructing image contents. Additionally, we develop a wavelet-based noise estimating network to improve the frequency perception and speed up the inference procedure. In future work, our focus will be on separating degraded images to extract more latent features that can better guide the image restoration process. We also aim to explore strategies to accelerate the reverse sampling process and create a more comprehensive dataset.

\bibliographystyle{IEEEtran}

\end{document}